\documentclass[acmtog,nonacm]{acmart}

\usepackage[mathletters]{ucs}
\usepackage[utf8x]{inputenc}
\usepackage{epsfig}
\usepackage{graphicx}
\usepackage{amsmath}

\usepackage{booktabs}
\usepackage{multirow}
\usepackage{bigdelim}
\usepackage{adjustbox}

\usepackage{geometrycollective} %

\newcommand{\MLP}{f_\theta}

\newcommand{\intlower}[1]{{#1}_{-}}
\newcommand{\intupper}[1]{{#1}_{+}}
\newcommand{\affine}[1]{\mathbf{\hat{#1}}}
\newcommand{\range}[1]{\textrm{range}({#1})}
\newcommand{\typepos}{\texttt{POSITIVE}}
\newcommand{\typeneg}{\texttt{NEGATIVE}}
\newcommand{\typeunk}{\texttt{UNKNOWN}}

\newcounter{algo}
\newenvironment{algo}[1]
{ \refstepcounter{algo}\noindent\rule{\columnwidth}{1.25pt}\vspace{-.2\baselineskip} \\ \textbf{Procedure~\thealgo} #1\vspace{-.55\baselineskip} \\ \noindent\rule{\columnwidth}{.5pt}\vspace{-1.2\baselineskip} }
{ \vspace{-.8\baselineskip}\noindent\rule{\columnwidth}{.5pt}\vspace{-\baselineskip} }

\algnewcommand{\LeftComment}[1]{\textcolor{commentblue}{\(\triangleright\)\textit{#1}}}

\setcopyright{rightsretained}
\acmJournal{TOG}
\acmYear{2022}\acmVolume{41}\acmNumber{4}\acmArticle{107}\acmMonth{7}\acmDOI{10.1145/3528223.3530155}

\acmSubmissionID{568}

\begin{document}

\title{Spelunking the Deep: Guaranteed Queries on General~Neural~Implicit~Surfaces~via~Range~Analysis}

\author{Nicholas Sharp}
\affiliation{%
  \institution{University of Toronto}
  \country{Canada}
}
\email{nsharp@cs.toronto.edu}

\author{Alec Jacobson}
\affiliation{%
  \institution{University of Toronto, Adobe Research}
  \country{Canada}
}
\email{jacobson@cs.toronto.edu}

\renewcommand{\shortauthors}{Sharp and Jacobson}

\begin{abstract}
\vspace{2em}
Neural implicit representations, which encode a surface as the level set of a neural network applied to spatial coordinates, have proven to be remarkably effective for optimizing, compressing, and generating 3D geometry.
Although these representations are easy to fit, it is not clear how to best evaluate geometric queries on the shape, such as intersecting against a ray or finding a closest point.
The predominant approach is to encourage the network to have a signed distance property.
However, this property typically holds only approximately, leading to robustness issues, and holds only at the conclusion of training, inhibiting the use of queries in loss functions.
Instead, this work presents a new approach to perform queries directly on \emph{general} neural implicit functions for a wide range of existing architectures.
Our key tool is the application of range analysis to neural networks, using automatic arithmetic rules to bound the output of a network over a region; we conduct a study of range analysis on neural networks, and identify variants of affine arithmetic which are highly effective.
We use the resulting bounds to develop geometric queries including ray casting, intersection testing, constructing spatial hierarchies, fast mesh extraction, closest-point evaluation, evaluating bulk properties, and more.
Our queries can be efficiently evaluated on GPUs, and offer concrete accuracy guarantees even on randomly-initialized networks, enabling their use in training objectives and beyond.
We also show a preliminary application to inverse rendering.
\vspace{2em}
\end{abstract}



\maketitle

\begin{figure}
\begin{center}
    \includegraphics[width=\columnwidth]{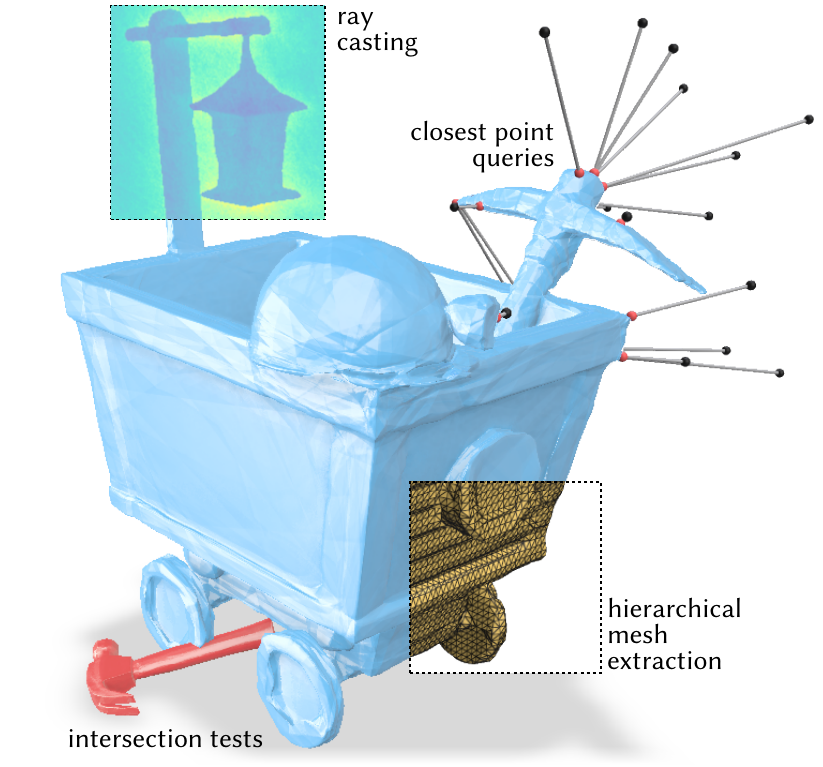}
    \caption{
      Our method enables geometric queries on neural implicit surfaces, without relying on fitting a signed distance function. 
      Several queries are shown here on a neural implicit occupancy function encoding a mine cart.
      These operations open up new explorations of deep implicit surfaces.
      \label{fig:teaser}
      \vspace{2em}
    }
\end{center}
\end{figure}

\vspace{2em}

\section{Introduction}

Representing shapes presents a fundamental dilemma across visual and scientific computing:
point clouds and voxel grids are easy to process efficiently, but lack explicit connectivity information; meshes offer a concise and precise description of a surface, but may require difficult unstructured computation, \etc{}
Recently, neural implicit representations have emerged as a promising alternative for a variety of important tasks---the basic idea is to encode a surface as a level set of a neural network applied to spatial coordinates.
These neural implicit surfaces inherit many of the strengths which have made neural networks ubiquitous across visual computing, including effective gradient-based optimization, integration with data-driven priors and objectives, and straightforward parallelization on modern hardware.

However, there is a price to pay in return for these strong properties: there is no clear strategy for evaluating even the most basic geometric queries against a neural implicit surface, such as intersecting a ray with the surface, or finding a closest point.
It would seem that the \emph{only} thing we can do with such a function is to sample it at a point.
In a sense, the powerful generality of neural networks is exactly what makes them difficult to query---because they can approximate arbitrary functions with adaptive spatial resolution, it is very difficult to characterize the geometry of their level sets.

One popular recourse is to attempt to fit implicit functions which not only encode a surface via their zero level set, but furthermore have a signed-distance function (SDF) property away from the level set: the magnitude of the function gives the distance to the surface.
Although exact SDFs are well-suited for many queries in geometry processing, approximate neural SDFs leave much to be desired.
First, such networks are only \emph{approximately} SDFs, and may overestimate the distance to the surface, causing queries to fail unpredictably.
More importantly, the SDF property only applies \emph{after} a network has been successfully fitted; thus we cannot make use of geometric queries in the early stages of training, \eg{}, to define geometric loss functions.
Even more broadly, relaxing the expectation that a network fits an SDF opens up a broader class of neural network formulations and objectives, such as those based on occupancy (\eg{}, as in \secref{InverseRendering}).

This work develops and studies a technique for performing queries on \emph{general} neural implicit surfaces—including not only SDFs, but also other functions which lack any special properties away from the zero level set.
Importantly, we do not define any new architectures or loss functions, but rather show how to perform queries on a broad class of existing networks, making our method immediately compatible with a wide range of past and future work on neural implicit formulations.
The result is a collection of subroutines for performing geometric queries on neural implicit surfaces, resolving a key weakness of the formulation and enabling promising new avenues of research.

To enable these queries, we leverage \emph{range analysis}, a class of automatic arithmetic techniques for computing bounds on the range of a function over a specified input domain.
These techniques were first popularized to bound the error incurred by floating point arithmetic, but can be applied more generally over any domain.
However, there exists a wide variety of range analysis methods, and these have not been previously studied in the context of neural implicit shapes.
In fact, we observe that many of these variants are entirely ineffective in this context, and a key component of this work is an investigation and empirical benchmark to identify range analysis schemes which are both efficient and effective for general neural implicit surface queries
(\secref{RangeAnalysis}).

\begin{figure}[b]
\begin{center}
    \includegraphics[width=\columnwidth]{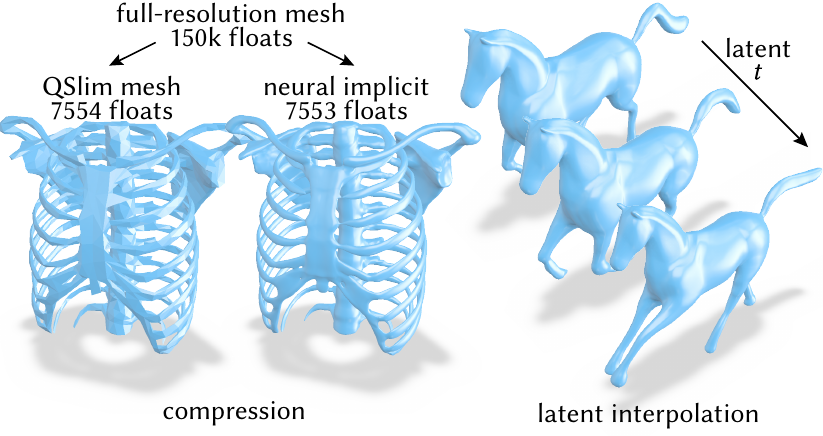}
    \caption{
      Neural implicit surface representations offer a variety of exciting properties, including efficient compression via gradient-based optimization (\figloc{left}), and interpolation through latent parameter spaces (\figloc{right}).
      Both examples are rendered in our ray casting framework.
      \label{fig:implicits_are_cool}
    }
\end{center}
\end{figure}

With range analysis of neural networks in hand as our primary tool, we develop a suite of geometric queries including ray casting, empty sphere queries, fast hierarchical surface sampling and mesh extraction, closest point queries, intersection tests, and more (\secref{GeometricQueries}). 
Many of these queries were not previously possible on general neural implicit surfaces, or could be evaluated only by dense brute-force sampling.
Our queries are guaranteed in the sense that they have bounded error with respect to the implicit surface regardless of the nature of the underlying neural network, \ie{} they apply even on a randomly initialized networks.
We demonstrate the potential of these queries applied to a wide variety of
problems in computer graphics, vision, and simulation.

\section{Background and Related Work}
\label{sec:Background}

The general literature on neural fields has rapidly exploded beyond the scope of this section, we point to the thorough survey of \citet{xie2021neural} as a general introduction.
In this work, we particularly consider queries on implicit \emph{solid surfaces}, as opposed to volumetric, partial occupancy, or participating media queries.
In particular, casting rays against solid surfaces has been very widely studied in computer graphics, and will serve as a proxy for many concerns that arise in the other queries we consider.

\begin{figure}[b]
\begin{center}
    \includegraphics[width=\columnwidth]{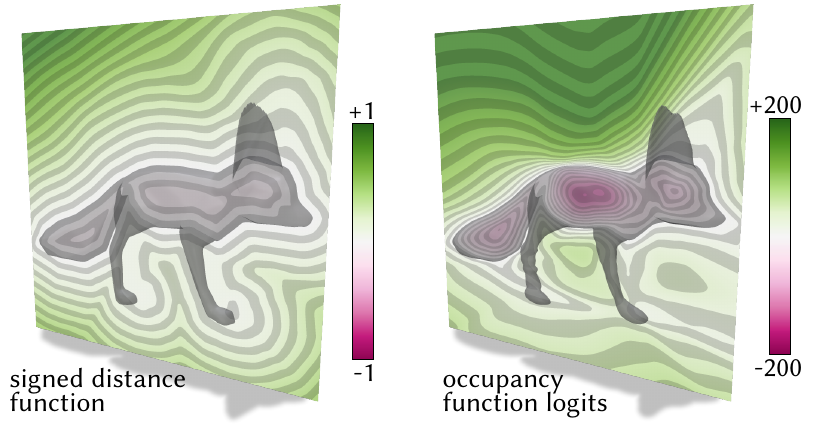}
    \caption{
      Neural implicit functions may fit a metric signed distance function (\figloc{left}), merely classify as an occupancy function (\figloc{right}), or use other formulations.
      The queries developed in this work apply in all cases. For occupancy, we plot the ``logits'' $y$ such that $\textrm{tanh}(y)$ gives occupancy.
      \label{fig:sdf_vs_occupancy}
    }
\end{center}
\end{figure}

\paragraph{Neural Implicit Surfaces}

\renewcommand{\SS}{\mathcal{S}} Neural implicit surfaces define the boundary 
$\SS$ of a solid 3D shape as the zero level set of a Multi-Layer Perceptron
(MLP) with internal parameters $θ$.
The function $\MLP$ takes as input a 3D position $x∈ℝ^3$, and possibly additional latent inputs, and outputs a scalar value:
\begin{equation}
\SS := \left\{ x∈ℝ^3 \left.\right| \MLP(x) = 0\right\}.
\end{equation} 
Neural implicit surfaces are immediately attractive because the \emph{sign} of their
forward evaluation \emph{classifies} $x$ as inside or outside the solid bounded
by $\SS$, typically assumed negative inside by convention.
Networks which instead model an \emph{occupancy} or \emph{density} (e.g., \cite{mescheder2019occupancy,ChenZ19}), such as the
geometric component of recently-popular NeRF models \cite{MildenhallSTBRN20},
can be viewed as an implicit surface by selecting an appropriate level set; this
connection was explored \citet{Yariv2021}.
More broadly, neural implicit surfaces are a special case of the larger family of
implicit surfaces defined by any arbitrary function $f(x)$, which have a vast
history in computer graphics (see, \eg{}, \cite{Bloomenthalbook,Menonbook}).
Classic implicit surfaces are constructed via trees of constructive
solid geometry operations with analytic functions at leaf nodes
(planes, spheres, cones, \etc{}) \cite{Ricci73}, or by crafting smooth blending
operations on radial basis functions (\eg{}, metaballs, blobs) \cite{Blinn82,WyvillMW86}.

\vfill

\begin{figure}
\begin{center}
    \includegraphics[width=\columnwidth]{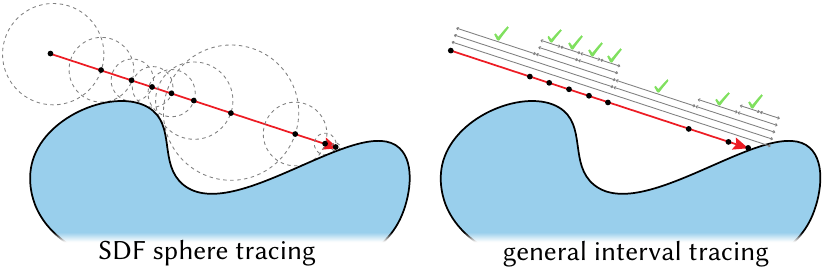}
    \caption{
      Sphere tracing finds the intersection of a ray with a shape by using a distance function to step forward (\figloc{left}).
      When a distance function is not available, but we can instead verify that intervals do not overlap the surface, interval tracing serves the same purpose (\figloc{right}).
      \label{fig:raycast_sdf_vs_general}
    }
\end{center}
\end{figure}

\pagebreak
\paragraph{Signed Distance Functions}
An important special case of implicit surfaces are \emph{signed distance functions} (SDFs), which add the requirement that the magnitude $|\text{SDF}(x)|$ is the distance to the closest point on the levelset $\SS$.
The SDF property implies that no point on the surface is within a distance $|SDF(x)|$ in any direction.
This observation forms the essence of the sphere tracing algorithm (\figref{raycast_sdf_vs_general}) for rendering implicit surfaces by casting rays
\cite{Hart1996Sphere,Balint2018Accelerating,Keinert2013Improved,Reiner2011Interactive}.

Often, a function need not satisfy the SDF property exactly, but merely be a \emph{weak SDF}, such that $|f(x)| < |SDF(x)|$.
A weak SDF still guarantees that no point on the surface is within a distance $|f(x)|$, which is sufficient for algorithms like sphere tracing to be correct, though they must take smaller steps.
Any smooth, bounded implicit function can be transformed to a weak SDF after scaling by its \emph{Lipschitz constant} $L$, though this is not necessarily productive in practice (see \secref{AlternativeApproaches}).
Representing arbitrary surfaces with exact SDFs is often difficult or unwieldy, but weak SDFs have 
been constructed to model fascinatingly complex surfaces (\eg{} by \citet{Inigo2008}).
Space-warped SDFs may no longer maintain a tight SDF, but \citet{SeybJNJ19} show that the inverse-warp function can afford sphere-tracing along curved rays in the unwarped domain.

Stated in the language of MLPs, if $\MLP$ is an SDF, or some appropriate Lipschitz constant $L$ is known \emph{a priori}, then a single forward evaluation would simultaneously reveal the implicit's
value at the current point along a ray \emph{and} the safe stepping distance along a cast ray (see \figref{raycast_sdf_vs_general}).
However, a generic neural implicit $\MLP$ will not automatically encode an SDF, nor have a known or small Lipschitz constant $L$.
The most widespread remedy in practice is to supervise training of the neural implicit function with precomputed SDF samples from known shapes.
If a network is well-trained to fit an SDF \cite{park2019deepsdf,Davies2020}, then sphere-tracing and other queries may be applied, though still with some risk of overzealous step scaling resulting in missed ray hits.

Another approach is by changing an MLP architecture to have a determinable global Lipschitz constant \cite{Yariv2021}. 
However, this may degrade surface fidelity, and moreover the global constant may be very high despite
being small in a region of interest (\eg{}, near $\SS$).
Computing precise local Lipschitz constants for common MLP architectures is
NP-Hard \cite{JordanD20,VirmauxS18}.

Yet another route is to incorporate loss functions encouraging Eikonality $\|∇\text{MLP}_θ\|
≈ 1$ \cite{atzmon2020sal,atzmon2020sald,GroppYHAL20,Davies2020} or bounded
Lipschitz constant \cite{Elsner2021}.
This approach may also effect surface fidelity 
and—even if successful—may only be true when training has completed,
precluding safe sphere-tracing type queries during training.

\paragraph{Querying General Implicit Surfaces}

What if we do not want to change the architecture of our $\text{MLP}$, or its training loss, to accommodate queries?
What if we are handed an $\text{MLP}$ which is not SDF-like?
For ray-casting, we could march with very small fixed steps \cite{PerlinH89}, but small steps are excessively expensive while large steps will cause rays to erroneously miss the surface (\figref{trace_comparison}).
We could contour the level set to a triangle mesh \cite{park2019deepsdf,GenovaCSSF20} via a
method such as marching cubes \cite{lorensen1987marching}, but this introduces
discretization error and aliasing, and also complicates differentiability
\cite{liao2018deep,RemelliLRGBBF20}.
For ReLU activations, the level-set will be polyhedral and could be \emph{theoretically} triangulated exactly, at significant cost \cite{LeiJ20}.
Even beyond ray-casting, we would seek many geometric queries to support the burgeoning geometry processing of neural implicit surfaces~\cite{yang2021geometry,yifan2021geometry}.

Instead, we look beyond the neural world toward past efforts of ray casting with a larger class
of arbitrary implicit functions.
A major theme of these works—dating at least to \citet{Duff1992}—is
to apply \emph{interval arithmetic} or more generally \emph{range analysis} with
its numerous variants \cite{rump2015implementation}.
Interval arithmetic is a code transformation technique to compute strict upper
and lower bounds for a composition of simple functions (see, \eg{},
\cite{stol1997self,moore2009introduction, alefeld2000interval} and our more
detailed discussion in \secref{RangeAnalysis}).

Applied to ray casting, range analysis can be employed to safely increase or
decrease the step size of a ray or bundle of rays 
\cite{HeidrichSS98,HeidrichS98,Knoll2007,KnollHKSHH09,mitchell1990robust,GalinGPP20,Keeter2020,FryazinovPC10,gamito2007ray,de1999interval,TBSCB:2021:TFDM},
or for other queries, such as finding closest points \cite{chan2008static}.
Despite this recurring interest, range analysis has not yet
been studied in the context of neural implicits to the best of our
knowledge.
In this paper, we demonstrate that carefully-selected variants of range analysis offer a highly effective strategy for analyzing general neural implicit surfaces, and show how a wide variety of geometric queries can be built upon it.

\section{Range Analysis of Neural Implicit Shapes}
\label{sec:RangeAnalysis}

We propose to apply \emph{range analysis} to neural implicit shape functions---these techniques take bounds on the input to a function, then apply automatic arithmetic rules to propagate bounds for each intermediate operation in a computation, and ultimately produce a bound on the function output~\cite{stol1997self}.
In our setting, range analysis can concretely bound the implicit function away from $0$ in a region, classifying the region as strictly inside or outside the shape.
This will be our foothold from which to build higher-level geometric queries (\secref{GeometricQueries}).

Range analysis has been widely studied within numerical computing, and particularly in the context of computer graphics (see \secref{Background}).
In the neural network literature, it has been leveraged for robustness and verification \cite{mirman2021fundamental,betancourt2021interval,gowal2018effectiveness,adam2016bounding,sahoo2015interval,dai2021lyapunov}, and even in the 3D setting by \citet{proszewska2021hypercube}, albeit in a voxel context.
However, the combination of these lines of research has not previously been realized by applying range analysis to neural implicit surfaces---we find this application to be very fruitful, but not without challenges.
For example, we observe that \emph{interval arithmetic}, the most basic and common range analysis, is largely ineffective when applied to our neural implicit surfaces.
For this reason, in this section we review core ideas in several forms of range analysis, and discuss particular issues which arise in application to MLPs, as well as performing an empirical study.
Ultimately, in \secref{SelectingAStrategy} we give concrete recommendations for variants of \emph{affine arithmetic} which are highly effective for range analysis of neural implicit surfaces.

\subsection{Interval Arithmetic}
\label{sec:IntervalArithmetic}

\begin{table*}
  \caption{We investigate several variants of range analysis for neural implicit surfaces, measuring the computation time relative to an ordinary scalar evaluation of the network (\emph{time}, lower is better), and the tightness of the bounds via the typical size of a region which can be bounded away from the zero by the analysis (\emph{length}/\emph{volume}, higher is better). 
  Both factors affect the efficiency of queries, exemplified via the time to cast rays, normalized by the fastest method (\emph{raycast}, lower is better).
  The last column summarizes recommendations from our study.
  See \appref{AdditionalDetails} for additional details.
  \label{tab:range_variants}
  }
\begin{tabular}{@{}rrcrcrl@{}}
\toprule
  \multicolumn{1}{l}{}               & \multicolumn{2}{c}{\textbf{analyze 1d region}}     & \multicolumn{2}{c}{\textbf{analyze 3d region}} & \multicolumn{1}{c}{\textbf{raycast}}   \\
  \textbf{Variant}                   & \quad time $\downarrow$ & length $\uparrow$          & \quad time $\downarrow$ & volume $\uparrow$      & time $\downarrow$ & \textbf{Comments} \\
\midrule
          interval &  2.4$\times$ &   0.011 &  2.3$\times$ &           < 0.001$\times 10^{-3}$ & 34.3$\times$ & \LeftComment{not effective, bounds too pessimistic} \\
     affine (full) & 95.3$\times$ &   0.821 & 96.4$\times$ & \hphantom{<}3.499$\times 10^{-3}$ &  8.4$\times$ & \LeftComment{\textbf{best for most volumetric queries on most networks}} \\
    affine (fixed) &  4.7$\times$ &   0.306 &  6.7$\times$ & \hphantom{<}0.267$\times 10^{-3}$ &  1.0$\times$ & \LeftComment{\textbf{best for ray casting on most networks}} \\
 affine (truncate) & 70.9$\times$ &   0.513 & 69.0$\times$ & \hphantom{<}0.906$\times 10^{-3}$ &  9.7$\times$ & \LeftComment{\textbf{best scaling to very large networks}} \\
   affine (append) & 29.3$\times$ &   0.351 & 31.0$\times$ & \hphantom{<}0.664$\times 10^{-3}$ & 58.3$\times$ & \LeftComment{no advantage \vs{} fixed/full} \\
    slope interval &  3.8$\times$ &   0.165 &  8.6$\times$ & \hphantom{<}0.042$\times 10^{-3}$ &  2.2$\times$ & \LeftComment{no advantage \vs{} affine}  \\
\bottomrule
\vspace*{.1em}
\end{tabular}
\end{table*}

Interval arithmetic~\cite{young1931algebra} is a technique for automatically computing bounds on the value of a function over a domain.
The basic idea is to replace each scalar quantity $x$ in a computation with a pair of bounds $[\intlower{x}, \intupper{x}]$.
Given bounds on the input $x$ to a function $y = f(x)$, arithmetic rules are applied to propagate the bounds forward through each elementary operation, eventually yielding bounds on the output $[\intlower{y}, \intupper{y}]$ such that $f(x) \in [\intlower{y}, \intupper{y}] \quad \forall x \in [\intlower{x}, \intupper{x}]$.
For example, elementary rules for addition, scalar multiplication, and the exponential are given by
\begin{align*}
  [\intlower{x}, \intupper{x}] + [\intlower{y}, \intupper{y}] &= [\intlower{x} + \intlower{y}, \intupper{x} + \intupper{y}]\\
  a [\intlower{x}, \intupper{x}] &= [\min(a\intlower{x}, a\intupper{x}), \max(a\intlower{x}, a\intupper{x})]\\
  \exp([\intlower{x}, \intupper{x}]) &= [\exp(\intlower{x}), \exp(\intupper{x})].
\end{align*}
In general, these rules can derived as needed, or looked up in a standard reference (\eg{} \cite{stol1997self}).
This technique extends directly to vector-valued quantities by tracking intervals for each component, and there are no restrictions on \eg{} the smoothness of $f$, so long as interval bounds can be derived for each constitutive operation.

These interval arithmetic rules already allow us to compute bounds on the output of an MLP.
However, it turns out that interval arithmetic alone is not a practical tool in our setting.

\paragraph{The Dependency Problem}
The main downside of interval arithmetic is that the computed bounds may be extremely pessimistic.
As an example, consider the simple operation $y \gets 2x - x$, evaluated on the range $x \in [-1,1]$. 
Clearly the actual bound on $y$ is $[-1,1]$, but applying the rules above yields a looser bound of $[-1,2]$ even in this simple example.
This issue is that the same interval-bounded quantity appears multiple times in the expression, and should cancel out, but the rules naively treat all interval quantities as being distinct.
In fact, the tightness of the bounds even depends on how the function is written algebraically---an unfortunate property in contrast to other automatic transformations such as automatic differentiation.
This effect is particularly problematic in MLPs, where linear layers $y \gets Ax$ involve a great deal of cancellation which is not captured by basic interval arithmetic, leading to extremely pessimistic bounds.
For this reason, we turn to an extension of interval arithmetic which tracks additional data to compute tighter bounds.

\begin{figure}
\begin{center}
    \includegraphics[width=\columnwidth]{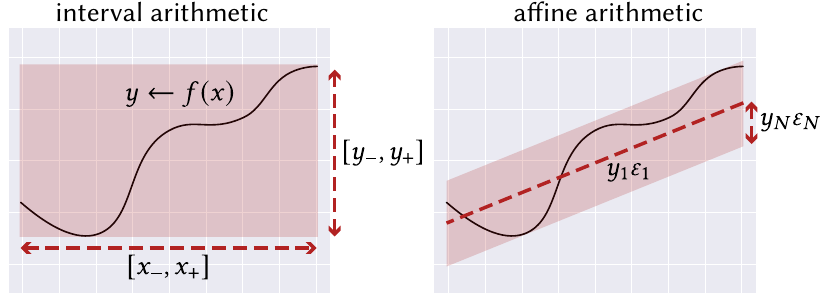}
    \caption{
      Affine arithmetic models the correlation between quantities to achieve much tighter bounds than interval arithmetic.
      \label{fig:interval_vs_affine}
    }
\end{center}
\end{figure}

\subsection{Affine Arithmetic}
\label{sec:AffineArithmetic}

Affine arithmetic~\cite{comba1993affine} generalizes interval arithmetic by tracking a collection of affine symbols, and offers the advantageous property that bounds are preserved exactly under affine operations.
In affine arithmetic, each scalar value $x$ is expanded to a base $x_0$ and a collection of affine coefficients $\{x_1,...,x_N\}$
\begin{equation}
  \label{eq:affine_def}
  \affine{x} = x_0 + \sum_{i=1}^N x_i \varepsilon_i, \quad \varepsilon_i \in [-1,1]
\end{equation}
where each $\varepsilon_i$ is a ``noise symbol'' representing some variation or uncertainty in the value of $\affine{x}$.
Crucially, this representation distinguishes distinct sources of variation: if $\affine{x}$ and $\affine{y}$ both vary due to some $\varepsilon_i$, we know that that these are the same, correlated uncertainties, and can \eg{} allow them to cancel when subtracting $x - y$.

Notice that for any uncertain quantity represented in affine form, we can easily read off bounds on the value of that quantity as
\begin{equation}
  \label{eq:def_range}
  \range{\affine{x}} = [x_0 - r,x_0 + r], \quad r = \sum_{i=1}^N |x_i|.
\end{equation}

Much like interval bounds, affine bounds can be propagated by automatic rules as a computation proceeds.
For instance, addition amounts to simply summing the base and coefficients
\begin{equation*}
  \affine{x} + \affine{y} = x_0 + \sum_{i=1}^N x_i \varepsilon_i + y_0 + \sum_{i=1}^N y_i \varepsilon_i = (x_0 + y_0) +\sum_{i=1}^N (x_i + y_i) \varepsilon_i
\end{equation*}
and likewise multiplication by a constant $a$ is given by
\begin{equation*}
  a \affine{x} = a x_0 + \sum_{i=1}^N a x_i \varepsilon_i.
\end{equation*}
Both operations are exact; they do not introduce new uncertainty.

For nonlinear functions $\affine{y} \gets f(\affine{x})$ such as $\exp$, $\tanh$, \etc{}, there is a straightforward recipe to propagate affine bounds by leveraging a linear approximation $f(x) \approx \overline{f}(x) := \alpha x + \beta$ on $\range{\affine{x}}$.
Letting $\gamma$ be the maximum error of this approximation $\gamma = \max_{x \in \range{\affine{x}}} |f(x) - \overline{f}(x)|$, then affine bounds can be propagated through $f$ according to
\begin{equation}
\label{eq:affine_func_update}
  \affine{y} = f(\affine{x}) = \overline{f}(\affine{x}) = 
  \alpha x_0 + \beta + \sum_{i=1}^N \alpha x_i \varepsilon_i + \gamma \varepsilon_{N+1},
\end{equation}
where $\varepsilon_{N+1}$ introduces $\gamma$ as a new, additional affine coefficient which models the nonlinear variation of $f(x)$ and is carried forward in subsequent computation.
For a particular nonlinear $f(x)$, we must then define $\alpha, \beta, \gamma$ for any input domain $\range{\affine{x}}$.
This approximation can be manually derived as needed, or looked up in a standard reference (\eg{} \cite{stol1997self}).
\appref{AffineArithmeticRules} gives formulae for $\alpha$,$\beta$,$\gamma$ for common activation functions, and furthermore explicitly defines all affine arithmetic rules used in this work.

Here, we consider only MLP-like neural network computations, and thus sidestep difficult operations in fully-general affine arithmetic implementations, such as multiplication between two affine terms ~\cite{rump1999intlab}.

\subsection{Reduced Affine Arithmetic}
\label{sec:ReducedAffineArithmetic}

In full affine arithmetic, each nonlinear operation introduces a new affine coefficient (\eqref{affine_func_update}), gradually increasing computational cost.
In our setting, each network layer of width $W$ would add $W$ new coefficients due to nonlinearities, which must then be propagated forward.
In an 8-layer 32-width network, this means that before the final dense layer what is normally a $\mathbb{R}^{32}$ vector will be replaced with a collection of $224$ $\mathbb{R}^{32}$ vectors encoding affine coefficients, each of which must be propagated via matrix multiplication, resulting in more than a $200\times$ increase in computation.
Although this cost may be worthwhile, it is pragmatic to consider alternatives.

Rather than retaining all affine coefficients, one can periodically reduce, or ``condense'' to some smaller set of coefficients, decreasing computational cost at the expense of potentially missing opportunities for cancellation \cite[\S3.18.1]{stol1997self}, \cite{gamito2007ray}.
Concretely, condensation replaces some set of affine coefficients at indices $\mathcal{D} = \{i_0,...i_N\}$ with a single new coefficient holding the sum of their magnitudes
\begin{equation}
  \texttt{condense}(\affine{x}, \mathcal{D}) = x_0 + \sum_{i \not\in \mathcal{D}} x_i \varepsilon_i + \big( \sum_{i \in \mathcal{D}} |x_i| \big) \varepsilon_{N+1}.
\end{equation}
One still must decide when to condense, and which coefficients to keep.
We consider four policies:
\begin{itemize}
  \item \texttt{affine-full}: no condensation, retain all affine terms
  \item \texttt{affine-fixed}: retain only affine terms from the original input domain; immediately condense all others
  \item \texttt{affine-truncate}: retain the $n_\textrm{keep}$ largest-magnitude terms 
  \item \texttt{affine-append}: after each nonlinearity, append the $n_\textrm{append}$ largest-magnitude new affine terms and condense the rest
\end{itemize}
In all cases we must also retain one additional affine term to hold the condensed value.
The tradeoffs between these strategies are not clear \apriori{}, motivating an empirical approach (\secref{SelectingAStrategy}).

\subsection{Selecting a Range Analysis Strategy}
\label{sec:SelectingAStrategy}

The extensive literature on automatic arithmetic for range analysis leads to many variants which \emph{could} compute bounds on neural implicit functions. 
However, these approaches vary drastically in their computational cost, and the tightness of the resulting bounds.

\paragraph{Performance}
Analyzing neural networks goes hand-in-hand with vectorized, GPU-based computation, and the performance trade offs therein.
For instance, vectorization implies maintaining the same set of affine terms for all quantities, precluding sparse representations with different terms for each.
Additionally, from the outset we consider only methods for which range analysis matrix multiplication can be implemented as a sequence of ordinary fast matrix multiplication primitives, which are crucial for performance.
Likewise, large amounts of dense arithmetic (\eg{} in \texttt{affine-full}) may be surprisingly performant compared to sorting and irregular data access (\eg{} in \texttt{affine-truncate}).

\paragraph{Correctness}
We emphasize that all considered variants of range analysis always yield \emph{correct} bounds, in the sense the output of the function on the interval is necessarily contained within the computed bounds.
Even floating-point inaccuracy can be addressed via careful control of rounding modes~\cite{stol1997self}, although we do not find it necessary in this work.
However, although these bounds are always correct, they are not necessarily \emph{tight}, and some variants of range analysis discussed above yield dramatically tighter bounds than others.
The tightness of the bounds in-turn affects the efficiency of downstream algorithms.

\paragraph{Empirical Study}
We conduct an empirical study to analyze the trade offs of range analysis techniques for neural implicit surfaces, which is to our knowledge the first in the context of neural networks.
We construct a dataset of neural implicit shapes fit via a variety of strategies, and for each we measure the tightness of the resulting bounds, as well as the added computational burden of range arithmetic---details are in \appref{AdditionalDetails}.
We consider ordinary interval arithmetic (\secref{IntervalArithmetic}), several variants of affine arithmetic (\!\secrefs{AffineArithmetic,ReducedAffineArithmetic}), and also a slope-interval form akin to the method of \citet{ratz1996optimized}, which combines interval arithmetic and automatic differentiation to bound derivatives.
See \tabref{range_variants} for results. 
Interval arithmetic is the least expensive, but yields extremely pessimistic bounds, and the slope-interval method offers little beyond affine arithmetic.
The \texttt{full}, \texttt{fixed}, and \texttt{truncate} variants of affine arithmetic all have advantages, depending on the context.

\paragraph{Recommendations}

We always recommend the use of affine arithmetic as opposed to interval or other arithmetics.
With this approach, a single implementation can easily adjust the truncation policy based on the task at hand.
For 1d ray casting queries, \texttt{affine-fixed} arithmetic offers the best performance, due to its low cost.
For spatial 3d queries, we find that surprisingly \texttt{affine-full} arithmetic is generally the most effective---although it performs a great deal of arithmetic, the tight bounds enable queries to explore a much smaller region of space.
Lastly, for large networks (\eg{} those with $>1000$ total scalar nonlinearities), \texttt{affine-truncate} may be a valuable alternative to avoid the scaling issues of full affine arithmetic, though sorting affine terms for truncation incurs significant overhead.
We use \texttt{affine-fixed} for all ray casting queries and \texttt{affine-full} for volumetric queries, unless otherwise noted.

\subsection{Applying Range Analysis}

We have now identified variants of affine arithmetic which are well-suited to computing range bounds on neural implicit functions, providing a foothold to design geometric queries.
To be clear, we do not propose any new network architecture or training objectives, but instead enable these queries directly on a wide range of existing architectures.
The class of networks to which our method applies includes MLPs, and is trivially extended to other common architecture components such as residual connections or latent inputs.
In principle, it can be applied to any layer operation for which an affine bound can be derived.

\setlength{\columnsep}{0.5em}
\setlength{\intextsep}{0em}
\begin{wrapfigure}{r}{78pt}
  \vspace{0.2em}
  \includegraphics{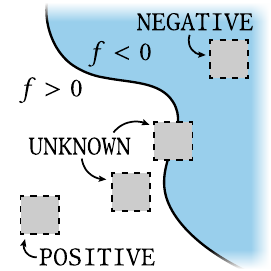}
\end{wrapfigure}
We will abstract over the use of range analysis via a function $\Proc{RangeBound}$, which takes as input an $s$-dimensional query box (which need not be axis-aligned), and classifies the value of the neural function as $\typepos$, $\typeneg$, or $\typeunk$ within the box.
For instance, ray casting requires bounds along a $1$d box which is not axis-aligned, while spatial queries decompose space into axis-aligned $3$d boxes.
Because the bounds from range analysis are not necessarily tight, $\Proc{RangeBound}$ may report $\typeunk$ over large queries, even if the function truly is bounded away from $0$; this can be resolved by subdividing and repeating the query over multiple smaller ranges.

This formulation is not limited to spatial coordinates; it applies to any network input such as latent parameters of a network, and the queries derived herein still apply with only small modifications.
Any network inputs for which we are not computing bounds are assigned the constant affine value $\affine{x} = x$.

\vspace*{4em} 

\pagebreak

\noindent
\begin{minipage}{\columnwidth} %
\vspace{2em}
\begin{algo}{\Proc{RangeBound}$(\MLP, c, \{v_i\})$}
  \label{alg:RangeBound}
  \begin{algorithmic}[1]
    \InputConditions{
      A function $\MLP : \mathbb{R}^d \to \mathbb{R}$ and a query box $B$ of dimension $s \leq d$ defined by its center $c \in \mathbb{R}^d$, and $s$ orthogonal box axis vectors $\{v_i \in \mathbb{R}^d\}$, not necessarily coordinate axis-aligned.}
    \OutputConditions{A bound on the sign of $\MLP(x) \enspace \forall x \in B$ as one of \\$\typepos$, $\typeneg$, or $\typeunk$.}
    \State {$\affine{x} \gets c + \sum_{i=1}^s v_i\varepsilon_i$}
    \Comment{Construct affine bounds defining the box}
    \State {$\affine{y} \gets \MLP(\affine{x})$}
    \Comment{Propagate affine bounds (\secref{AffineArithmetic})}
    \State{$[\intlower{y}, \intupper{y}] \gets \range{\affine{y}}$} \Comment{Bound the output (\eqref{def_range})}
    \If{$\intlower{y} > 0$} \Return \typepos \EndIf
    \If{$\intupper{y} < 0$} \Return \typeneg 
    \Else{ \Return \typeunk}
    \EndIf
  \end{algorithmic}
\end{algo}
\vspace{1em}
\end{minipage}

\subsection{Implementation}
\label{sec:Implementation}

To facilitate integration in deep learning pipelines, we implement affine arithmetic for neural implicit surfaces, as well as our geometric queries, in the JAX framework~\cite{jax2018github}.
We also investigated a prototype JAX range analysis implementation as a general transformation applied to arbitrary JAX programs---we ultimately found that specifying to MLPs resulted in a more efficient implementation, but this is an exciting avenue for future investigation.
Range analysis on neural networks and the queries below are evaluated entirely on the GPU, where we leverage parallel traversals and dynamic batching to compute efficiently with fixed-size array kernels; an implementation is included as supplementary material.
All timings are measured on an RTX 2070 GPU.

To ensure correctness, we also validate our range analysis bounds by fuzz-testing \algref{RangeBound} on $10^6$ randomly sampled input regions with a variety of network architectures, ensuring that point-sampled function evaluations always lie within floating point tolerance of the computed bounds.
These tests succeed in all cases.

\begin{figure}[b]
\begin{center}
    \includegraphics[width=\columnwidth]{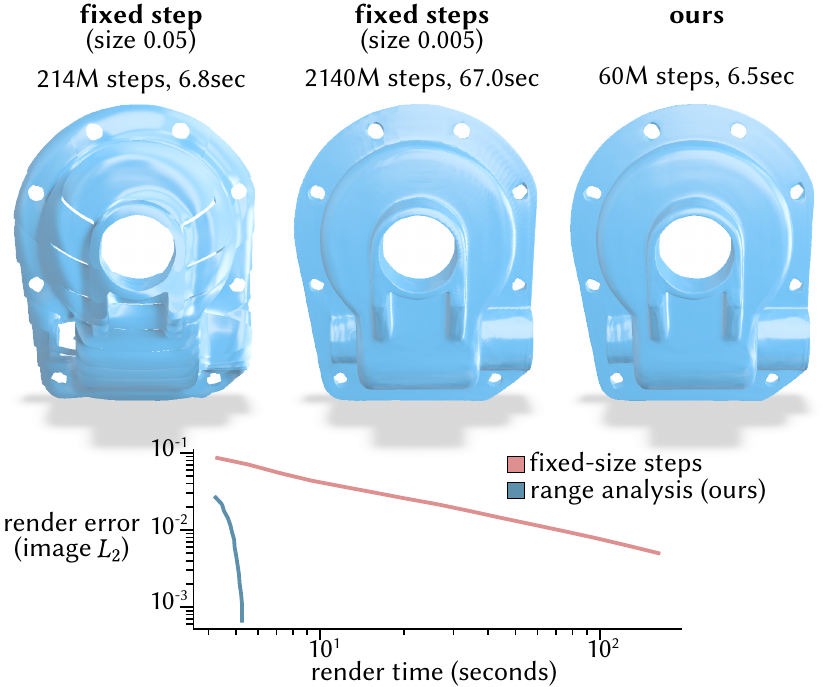}
    \caption{
      Previously, casting rays against general neural implicit surfaces required small, fixed-sized steps, used here to render an image.
      Large steps can miss parts of the surface (\figloc{top, left}), but smaller steps are expensive (\figloc{top, middle}).
      Our approach avoids tuning a step size, and is much faster. (\figloc{top, right}).
      We measure this effect via the image error at various render time budgets (\figloc{bottom}), varying the fixed step size or our convergence parameter $\delta$, respectively.
      \label{fig:trace_comparison}
    }
\end{center}
\end{figure}

\section{Geometric Queries}
\label{sec:GeometricQueries}

Range analysis via affine arithmetic now provides the key missing tool for geometric operations on general neural implicit surfaces by efficiently computing bounds on the value of the implicit function over spatial regions.
In this section, we develop a variety of useful geometric queries using these bounds---in many cases, these queries are possible on general neural implicit surfaces for the first time.
\appref{AdditionalDetails} lists additional configuration details for experiments.

\paragraph{Visualization}
All figures are rendered via direct ray casting of neural implicit surfaces using our ray cast query (\secref{RayCasting}), and are shaded via material capture (except in \figref{inverse_render_bunny}). Ground shadows are evaluated via casting rays upward followed by a Gaussian blur, and an ambient occlusion term is approximated by sampling points on a hemisphere, again both using our ray casting operation.

\vspace{4em}

\pagebreak
\subsection{Defining Convergence and Correctness}
\label{sec:Convergence}
Selecting an appropriate convergence criterion is a subtle but important dilemma in all of the queries we would like to perform, both to avoid excess computation, and to ensure termination even in imperfect floating-point arithmetic.
In other settings (\eg{} implicit SDFs), convergence can be defined in terms of the magnitude of the implicit function, when some $|f(p)| < \epsilon$.
However, such a convergence test is not appropriate for general implicit functions, where the magnitude of $f$ might vary wildly, and is not known \apriori{}.
Similarly, it is infeasible to provably capture all, pathologically small features which might exist in an implicit surface, given the bounded accuracy of numerical computation.

\setlength{\columnsep}{0.5em}
\setlength{\intextsep}{0em}
\begin{wrapfigure}{r}{78pt}
  \vspace{0.2em}
  \includegraphics{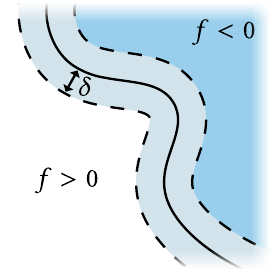}
\end{wrapfigure}
Instead, we argue for defining convergence and correctness in the sense that the output of a query must be correct for some surface which is a dilation or contraction of the true level set by at most some small $\delta$ (see inset). 
For example, in the case of ray casting below, this criterion implies detecting convergence whenever a point $p$ within $\delta$ along the ray has $f(p)$ with the opposite sign from the origin, and it also means that when not converged, a step of size $\delta$ is always safe.
Similar convergence policies are used in all other queries as well---we guarantee that results are within $\delta$ of the true level set, and that a region of the surface of size at least $\delta$ is never ``missed'' by the query, no matter what the magnitude of $f$ might be.
We use $\delta=0.001$ for all examples, on shapes normalized to the unit sphere.

\subsection{Ray Casting}
\label{sec:RayCasting}

Finding the intersection with a ray is the most common and widely-studied geometric query on implicit surfaces.
Precisely, given a ray source $p \in \mathbb{R}^3$ and direction $r \in \mathbb{R}^3$, we seek the smallest $t > 0$ such that $f(p + t r) = 0$.
Often these rays are primary pixel rays in some view of the scene, but ray casting also arises more broadly in rendering and beyond.

We adopt a simple strategy of marching along the ray, attempting steps of size $\sigma$: if range arithmetic can bound the implicit function away from $0$ on the step interval, we take the step and increase $\sigma$ by a factor $\eta_+$, otherwise we decrease $\sigma$ by a factor of $\eta_-$ and retry the step.
Similar approaches have a long history in computer graphics~\cite{mitchell1990robust,de1999interval,gamito2007ray}.
Equipped with range bounds on MLPs from \secref{RangeAnalysis}, we can now apply this strategy to general neural implicit surfaces for the first time.
This scheme is described precisely in \algref{CastRay}.
It is used to render all visualizations in this work, and \figref{trace_comparison} gives a comparison to alternative approaches.

Validating steps with range analysis guarantees correct output regardless of the choice of initial step size, and adapting the step size by $\eta_+,\eta_-$ automatically adjusts to an appropriate scale for the problem.
We suggest $\sigma_0 = {t_{\max}}/{10}$, $\eta_- = 0.5$, and $\eta_+=1.5$ as reasonable step size parameters, and use these values in all experiments.
The miss threshold $t_{\max}$ depends on the length scale and typical casting distance; we use $t_{\max} =10.$ for shapes normalized to the unit sphere.
In principle, ray casting queries could be further accelerated using the spatial bounding hierarchy from \secref{SpatialHierarchies}.

\noindent
\begin{minipage}{\columnwidth} %
\vspace{3em}
\begin{algo}{ \Proc{CastRay}$(\MLP, p, r)$ }
  \label{alg:CastRay}
  \begin{algorithmic}[1]
    \InputConditions{
      An implicit surface $\MLP : \mathbb{R}^d \to \mathbb{R}$, and a ray source and direction $p,r \in \mathbb{R}^3$.
    }
    \OutputConditions{
      The distance $t$ to the ray-surface intersection, or no hit.
    }
    \State {$\sigma \gets \sigma_0, \quad t \gets 0$} \Comment{Initialize steps}
    \State {$f_0 \gets \MLP(p)$} 
    \While {$t < t_{\max}$} \Comment{March forward}
      \State {$p_c \gets p + (t + \delta) r$} \Comment{Convergence test}
      \State {$f_c \gets \MLP(p_c)$} 
      \If{$\Proc{DifferentSigns}(f_0, f_c)$} \Return $t$ \Comment{Found hit} \EndIf
      \State {$c \gets p + (t + \sigma / 2)r$} 
      \Comment{Construct 1d query box}
      \State {$v \gets (\sigma / 2)r$}
      \If{$\Proc{RangeBound}(\MLP, c, \{v\}) \neq \typeunk$} \Comment{\secref{RangeAnalysis}}
        \State {$\sigma^* \gets \sigma, \quad \sigma \gets \sigma \eta_+$} \Comment{Step is safe, increase step size}
      \Else
        \State {$\sigma^* \gets 0, \quad \sigma \gets \sigma \eta_-$} \Comment{Step is not safe, decrease step size}
      \EndIf
      \State{$t \gets t + \max(\sigma^*, \delta)$} \Comment{Take step (tolerance $\delta$ is always safe)}
    \EndWhile
    \State {\Return \texttt{NO\_HIT}}
  \end{algorithmic} 
\end{algo} 
\vspace{1em}
\end{minipage}

\pagebreak
\subsection{Frustum Ray Casting}
\label{sec:FrustumRayCasting}

When casting primary rays to render an image, rays from adjacent pixels traverse similar regions of space, wastefully repeating similar computation---an effect which becomes more pronounced at higher resolutions.
This observation has led to a variety of techniques in traditional ray tracing and even range analysis which process groups of nearby rays simultaneously \eg{} by \citet{reshetov2005multi,florez2006improving}.
Similar concerns arise in neural volumetric rendering (\eg{} \cite{barron2021mip}), although there the objective is approximate anti-aliasing moreso than exact spatial acceleration.
By applying 3D range analysis to a box which bounds a frustum of rays, we can accelerate ray casting against existing neural implicit surfaces while still guaranteeing precisely correct results.

\begin{figure}
\begin{center}
    \includegraphics[width=\columnwidth]{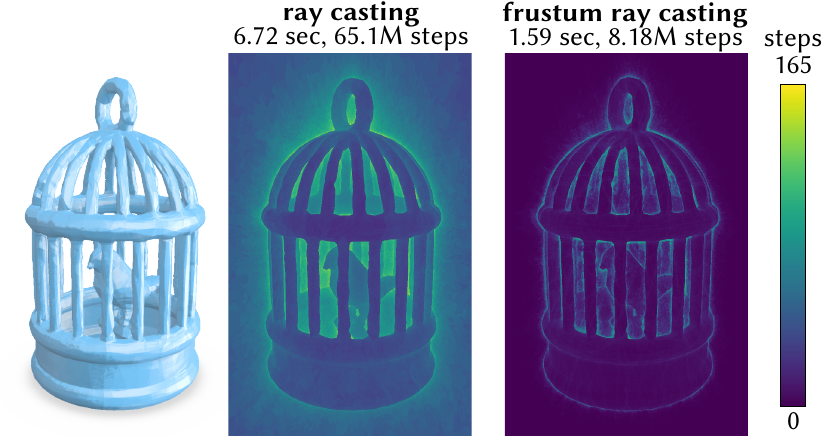}
    \caption{
      Range analysis enables ray casting of general neural implicit functions, here applied to a neural occupancy function.
      We can also use 3D range analysis to bound volumetric regions of space, enabling \emph{frustum ray casting} where blocks of rays are marched forward in a single evaluation, without any approximation error.
      Plots show the amortized number of marching steps per pixel; frustum steps are shared by many rays, greatly decreasing the total number of steps. Rendered at $1024 \times 1024$ resolution.
      \label{fig:trace_frustum_comparison_birdcage}
    }
\end{center}
\end{figure}

\setlength{\columnsep}{1em}
\setlength{\intextsep}{0em}
\begin{wrapfigure}{r}{123pt}
  \includegraphics{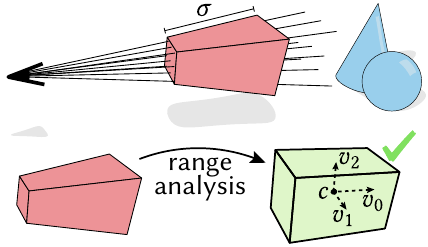}
\end{wrapfigure}
Our strategy is to initialize a coarse set of frusta over the primary pixel rays in an image, and march each frustum forward with steps similar to \algref{CastRay} (see inset).
The frusta are dynamically subdivided whenever the step size $\sigma$ becomes smaller than the width of the frustum, eventually reducing to individual pixel rays as they hit the surface.
Although 3D range analysis over boxes is moderately more expensive than 1D analysis  (see \tabref{range_variants}), the algorithmic advantage of amortizing locally-similar computation leads to significant performance improvements over casting individual rays (\figref{trace_frustum_comparison_birdcage}).
This gap widens as resolution increases, improving scaling for high-fidelity renderings.

\subsection{Empty Spheres}
\label{sec:EmptySpheres}

\setlength{\columnsep}{1em}
\setlength{\intextsep}{1em}
\begin{wrapfigure}{r}{110pt}
  \includegraphics{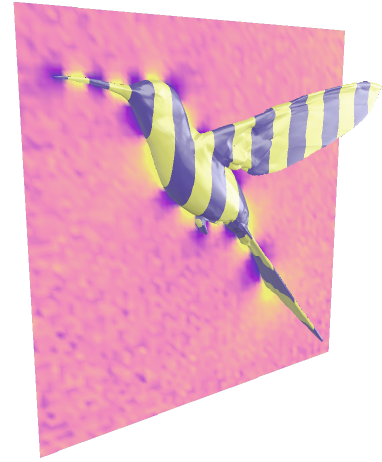}
  \caption{
    A slice of a solution to a scalar Laplace problem with Dirichlet boundary values on a neural implicit surface, approximated via random walks. 
    Range analysis enables empty-box queries with a throughput of $196$k queries per second.
    \label{fig:walk_on_spheres}
  }
\end{wrapfigure}

An \emph{empty sphere} query at a point $p$ reports the radius of a sphere at $p$ which does not intersect the surface.
This radius is not required to be maximal, though larger radii are preferred.
Much like interval ray marching, we can evaluate empty sphere queries by applying range analysis on a box centered at $p$: if range arithmetic determines the box is \typepos{} or \typeneg{} then we have answered the query, and if not we try a smaller box.

A natural application of empty sphere queries is traversing random walks in space, repeatedly moving to a random point on the empty sphere until eventually reaching the surface.
These random walks enable a grid-free Monte-Carlo scheme for the Poisson-like PDEs which are widespread in graphics and geometry processing \cite{Sawhney:2020:MCG,nabizadeh2021kelvin}.
However, the true value of bounding 3d spatial regions arises from constructing hierarchies over the domain.

\subsection{Spatial Hierarchies}
\label{sec:SpatialHierarchies}

Hierarchical spatial acceleration structures are a cornerstone tool in high-performance visual computing, from bounding volume hierarchies to mipmaps.
Recent work has likewise shown dramatic performance benefits in the neural implicit context (\eg{} \cite{takikawa2021neural,barron2021mip}).
However, these methods target fast forward evaluation and training rather than evaluating geometric queries as investigated here, and introduce new customized architectures.
Instead, we will show that range analysis can be used to build guaranteed hierarchies over arbitrary \emph{existing} neural implicit surfaces, even ordinary MLPs, enabling a variety of fast guaranteed spatial queries.

First, in \algref{BuildSpatialTree} we describe a general branch-and-bound procedure for constructing a bounding $k$-D tree of a neural implicit surface using range analysis.
The queries below in \secref{SurfaceSampling} through \secref{ClosestPoint} all make use of variants of this strategy, adapting either the refinement criterion for the tree or the manner in which it is traversed.

On convergence the resulting tree has guaranteed accuracy, in the sense that nodes classified as $\typepos$ or $\typeneg$ are necessarily strictly outside or inside of the shape, respectively, and $\typeunk$ nodes are within $\delta$ of the level set.
This property in-turn enables spatial queries with guaranteed accuracy by evaluation over the set of tree nodes.
An exception to the convergence policy is \secrefs{SurfaceSampling,HierarchicalMeshExtraction}, where the query demands refinement to a predefined depth.
In practice, \Proc{BuildSpatialTree} is not implemented recursively, but iteratively in parallel rounds of exploring and expanding batches of nodes.
A similar strategy could be used to build other acceleration structures such an octree or bounding volume hierarchy, but here we use a $k$-d tree because it is simple and extends trivially to higher dimension (\eg{} for bounding with respect to latent parameters).

In principle a similar tree could be constructed without range analysis, via a sufficiently dense grid of samples to classify nodes in the sense of ~\secref{Convergence}.
However, such an approach would be dramatically more expensive. 
As an example, constructing such a tree for the shape in \figref{extract_mesh_snake} would require 5.3B function samples, which already require 193 seconds to evaluate; building our tree takes just 0.610 seconds.

\noindent
\begin{minipage}{\columnwidth} %
\vspace{2em}
\begin{algo}{ \Proc{BuildSpatialTree}$(\MLP, x_l, x_u)$ }
  \label{alg:BuildSpatialTree}
  \begin{algorithmic}[1]
    \InputConditions{
      An implicit surface $\MLP : \mathbb{R}^d \to \mathbb{R}$, and domain bounds $x_l,x_u \in \mathbb{R}^d$ as lower and upper corners of bounding box.
    }
    \OutputConditions{
      A $k$-d tree bounding the level set of $\MLP$.
    }
    \State {\LeftComment{Classify the value of $\MLP$ in the node (\secref{RangeAnalysis})}}
    \State {$x_c \gets (x_l + x_u) / 2$}
    \State {$v \gets \Proc{Diag}(x_u - x_c)$} \Comment{Diagonal matrix from vector}
    \State {$t \gets \Proc{RangeBound}(\MLP, x_c, v)$}
    \State {}

    \State {\LeftComment{Test convergence for nodes away from the level set}}
    \If{$t \in \{\typeneg, \typepos\}$}
      \State{\Return $\{(x_l, x_u, t)\}$}
    \EndIf
    \State {}
    
    \State {\LeftComment{Test convergence for small nodes near surface (\secref{Convergence})}}
    \State {\LeftComment{(alternately, recurse to some fixed depth instead)}}
    \If{$\max(x_u-x_l) < \delta / \sqrt{d}$} 
      \State {$\{p_i\} \gets \Proc{PointOnEachFaceOfNode}(x_l, x_u)$}
      \If{$\mathrm{any}(\MLP(p_i) < 0) \land \mathrm{any}(\MLP(p_i) > 0)$}
        \State{\Return $\{(x_l, x_u, t)\}$} \Comment{Node is within $\delta$ of level set}
      \EndIf
    \EndIf
    \State {}
    
    \State {\LeftComment{Compute a split point}}
    \State {$i_s \gets \textrm{argmax}{(x_u - x_l)}$} \Comment{Widest dimension of node}
    \State {$x_s \gets 0^d, \quad x_{s,i_s} \gets (x_u - x_l)_{i_s}$} \Comment{Vector to new midpoint}

    \State {}
    \State {\LeftComment{Recurse on both subtrees and return the union of all nodes}}
    \State {$\mathcal{T}_a \gets \Proc{BuildSpatialTree}(\MLP, x_l, x_u - x_s)$}
    \State {$\mathcal{T}_b \gets \Proc{BuildSpatialTree}(\MLP, x_l + x_s, x_u)$}
    \State {\Return $\mathcal{T}_a \cup \mathcal{T}_b$}
  \end{algorithmic} 
\end{algo} 
\vspace{1em}
\end{minipage}

\subsection{Surface Sampling}
\label{sec:SurfaceSampling}

\setlength{\columnsep}{1em}
\setlength{\intextsep}{0em}
\begin{wrapfigure}{r}{130pt}
  \vspace{0.2em}
  \includegraphics{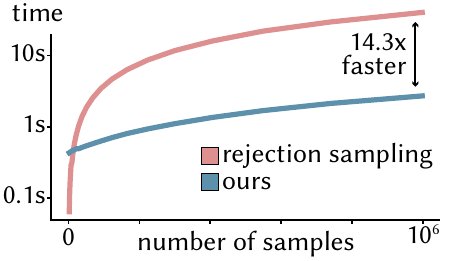}
  \vspace*{-2em}
  \caption{
    A runtime comparison of implicit surface sampling schemes.
    \label{fig:sampling_perf_plot}
  }
\end{wrapfigure}
Sampling a set of points on a surface is a common operation in geometric machine learning, often used to evaluate a loss function or metrics such as Chamfer distance.
Consider in particular sampling $N$ points which have $|f(p)| < r$.
The naive approach is a rejection strategy, uniformly sampling the domain until enough valid points are found.
Instead, our spatial hierarchy can be used to find a set of nodes which necessarily contain all regions with $|f(p)| < r$; sampling from these nodes only rather than the whole domain is dramatically more efficient.
We leverage the $k$-D tree described in \secref{SpatialHierarchies}, refining to a fixed depth while discarding nodes which are classified by range analysis as having $f > r$ or $f < -r$.

\begin{figure}
\begin{center}
    \includegraphics[width=\columnwidth]{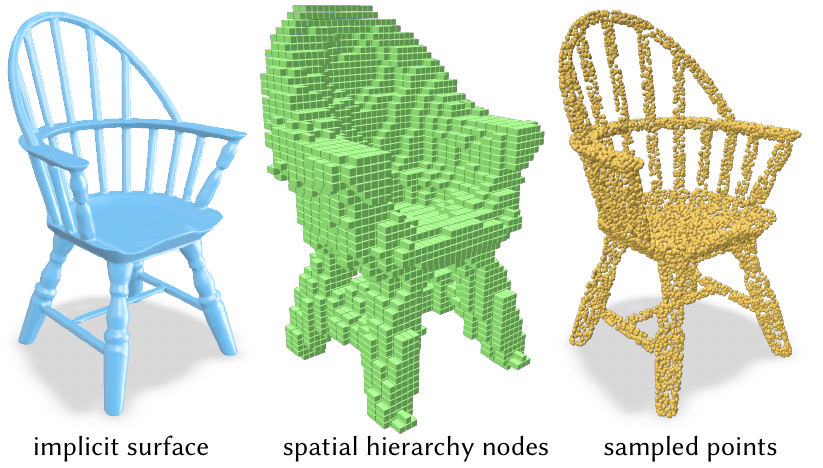}
    \caption{
      Sampling a points on a surface is a common task in geometric learning.
      Our spatial hierarchies can be used to sample efficiently, outperforming naive rejection sampling by an order of magnitude for large sample sets.
      \label{fig:sampling_chair}
    }
\end{center}
\end{figure}

\figref{sampling_chair} shows the result of this process, and \figref{sampling_perf_plot} plots the corresponding runtime for generating a specified number of samples with our method and with naive rejection sampling.
After an initial cost of building the hierarchy, our method becomes an order of
magnitude faster---this gap widens as smaller sampling bands are used or the
number of samples increases.

For applications requiring points lying more precisely on the surface, our sampling may benefit from ``polishing'' using the method of \citet{WangWOS21}, but this is left as future work.

\subsection{Hierarchical Mesh Extraction}
\label{sec:HierarchicalMeshExtraction}

Implicit representations are appealing for their mesh-free nature, but
nonetheless it is common to extract an explicit triangle mesh of the level set,
either as a format conversion or to enable operations defined only on a
mesh~\cite{park2019deepsdf}.
Our spatial hierarchy can be leveraged to accelerate marching cubes mesh extraction~\cite{lorensen1987marching} by only extracting from cells near the surface, while necessarily producing the same output as brute-force extraction.

In this case, given an extraction resolution $2^m$ along each dimension, we
build a spatial hierarchy via \algref{BuildSpatialTree} to a fixed depth of $3m$
while retaining all nodes classified as $\typeunk$.
The factor of $3$ arises from splitting the $k$-D along each dimension.
Applying marching cubes extraction individually in each node then generates the resulting mesh---the nodes excluded by the hierarchy necessarily would not contribute.
In practice, we apply dense extraction at the bottom $l=3$ levels of our hierarchical for performance.

The hierarchical nature of our mesh extraction bears resemblance to the method of \citet{mescheder2019occupancy}.
When the network is smooth, we suspect their method could be faster, though
surface features smaller than their initial \emph{low}-resolution grid could be missed.
In contrast, our method is guaranteed to split cells that contain the surface,
with error bounded by the \emph{finest} grid resolution. Moreover, our method
will work for arbitrarily misbehaving implicits (e.g., randomly initialized
networks, for use in loss functions).

\begin{figure}
\begin{center}
    \includegraphics[width=\columnwidth]{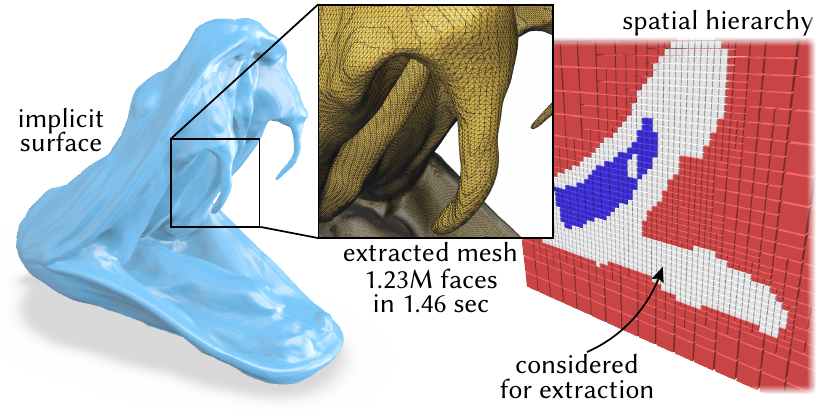}
    \caption{
      If an explicit mesh is desired, our spatial bounding hierarchy enables an adaptive variant of marching cubes which avoids evaluating the function in large empty regions.
      This yields the same output as ordinary dense marching cubes, but scales much more efficiently to high-resolution meshes.
      \label{fig:extract_mesh_snake}
    }
\end{center}
\end{figure}

\figref{extract_mesh_snake} shows the result of this procedure, where a $1.23$ million face mesh is extracted in $1.46$ seconds, including the time to build the hierarchy.
Merely evaluating $f$ at an equivalent grid of dense points would take $5.03$ seconds.
In fact, this procedure scales remarkably well to even higher resolution meshes: extracting a $4.9$ million face mesh of the same implicit surface takes just 4.13 seconds, while dense evaluation would require $39$ seconds. 
As an added bonus, the resulting mesh lies in the leaf nodes of a k-D tree by construction---it already has a spatial acceleration structure ready for subsequent processing if desired.

Here we treat only classic marching cubes, other approaches such as dual contouring~\cite{ju2002dual} could be applied using a similar strategy.
Additionally, our spatial acceleration is also compatible with differentiable
variants of mesh extraction~\cite{liao2018deep,shen2021deep}, though we do not
yet pursue an implementation.

\subsection{Bulk Properties}
\label{sec:BulkProperties}

\setlength{\columnsep}{0.75em}
\setlength{\intextsep}{0em}
\begin{wrapfigure}{r}{122pt}
  \vspace{0.2em}
  \includegraphics{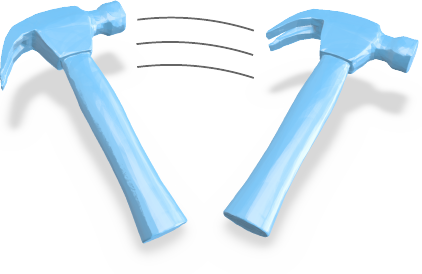}
\end{wrapfigure}
Physical simulation on implicit shapes requires evaluating bulk properties such as mass and moments of inertia via integrals over the interior of the shape.
The spatial hierarchy (\algref{BuildSpatialTree}) makes it straightforward to evaluate these integrals by accumulating contributions from all interior nodes.
We estimate the contribution from any $\typeunk$ nodes straddling the boundary via random sampling, akin to \secref{SurfaceSampling}.
If the tree is fully refined, the resulting values already satisfy the convergence guarantee in \secref{Convergence}, although faster runtimes can be obtained by refining to some fixed depth.
If desired, we can further bound the possible error in an integral via the largest and smallest possible contribution from any $\typeunk$ nodes in the hierarchy.
The mass of the inset shape is computed to a relative accuracy of $5\times10^{-4}$ via this strategy in $0.77$ seconds, \vs{} $6.72$ seconds for integration with uniform random samples.

\subsection{Intersection and Collisions}
\label{sec:IntersectionAndCollisions}

Given two neural implicit shapes, how can we test whether they intersect one another?
This basic operation will be increasingly necessary for tasks like path planning and simulation if neural implicit surfaces are to be incorporated in realistic virtual environments.
Our spatial hierarchy (\algref{BuildSpatialTree}) can be used to test for intersections by simultaneously subdividing the tree with respect to two implicit functions.
If either function is bounded $\typepos$ in a node, then that node necessarily does not contain an intersection.
Repeatedly subdividing the tree either yields a node in which the surfaces intersect, or verifies that there is no such intersection.

\begin{figure}
\begin{center}
    \includegraphics[width=\columnwidth]{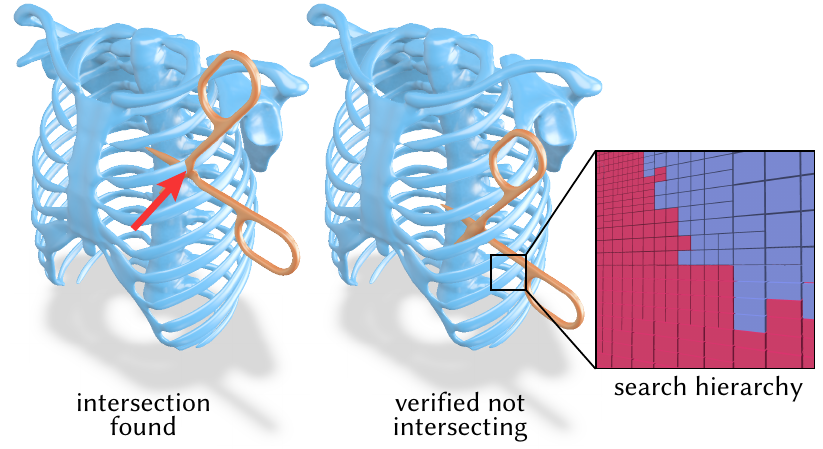}
    \caption{
      Our queries are used to detect intersections between a pair of general neural implicit shapes.
      The inset image shows the k-d tree used to bound the space.
      Each cell has been verified to not overlap with at least one of the two shapes, indicated by the color.
      \label{fig:intersection_ribcage}
    }
\end{center}
\end{figure}

\figref{intersection_ribcage} shows this procedure testing for intersections between two neural implicit surfaces, one of which is encoded as an SDF and the other as an occupancy function.
The runtime for that example is 80ms per query; computing the query to the same accuracy guarantee via a densely sampled grid would require 5.3B function samples and several minutes of processing.
We note that, coupled with the bulk properties in ~\secref{BulkProperties}, we have now developed the core computational ingredients for rigid body simulation of neural implicit surfaces, an exciting avenue for ongoing applications.

\subsection{Closest Points}
\label{sec:ClosestPoint}

Given a point in space, a \emph{closest point query} seeks the nearest location on the neural implicit level set.
On an exact signed distance function the nearest point on the surface can be computed as $p_\textrm{nearest} = p - f(p) \nabla f(p)$, but for approximate neural SDFs and more general neural implicit surfaces, there is no such clear strategy.
Fortunately, our hierarchical bounding $k$-d tree is also a natural data structure to perform closest point queries.
Indeed, finding nearest-neighbor points is a classic application of $k$-d trees; the only nuance in this case is that the target set of points is a continuum encoded by the implicit function.

\begin{figure}
\begin{center}
    \includegraphics[width=\columnwidth]{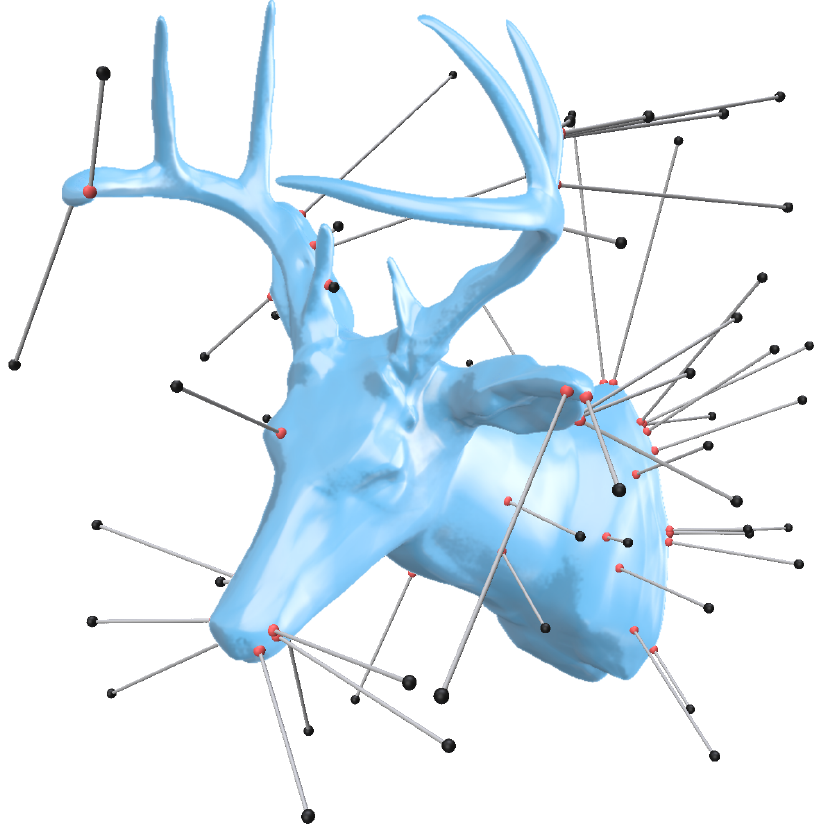}
  \caption{
    We project query points to the closest point on the neural implicit level set.
    Black query points are sampled randomly in space, red points are their projections.
    \label{fig:closest_point_deer}
  }
\end{center}
\end{figure}

Given a query point $p$, we descend the tree from \algref{BuildSpatialTree}: if test points sampled on the faces of a node include both signs of the implicit function, then the node necessarily spans the surface, and we update the closest-point distance as the farthest distance from $p$ to any point in the node, using the node center as the corresponding location.
Nodes which are bounded away from the level set need not be explored, as well as those for which the nearest point in the node is farther than the closest distance already found.
As elsewhere, the near-surface convergence criterion described in \algref{BuildSpatialTree} ensures a result within a small $\delta$ of the true level set (\secref{Convergence}).
\figref{closest_point_deer} shows the result of this procedure; each closest point query takes $70$ms on average.
Evaluating the same queries by expanding a dense grid samples would require on-average $604$M function evaluations, for a runtime of 22.2 seconds per query.
In this setting, unlike most other volumetric queries, we find \texttt{affine-fixed} evaluation to be about $3\times$ faster than \texttt{affine-full}, perhaps because quickly discovering a reasonably-near point truncates the search more effectively than tighter range bounds.
Our implementation is designed to minimize latency: each query point performs an independent lazy traversal of the tree, usually only exploring a small subset of the possible nodes.
If desired, the entire tree could instead be constructed initially and then directly traversed; which would decrease the cost per query at the expense of some initial precomputation.

\subsection{Alternative Approaches}
\label{sec:AlternativeApproaches}

In general, there are not well-established prior strategies for geometric queries on general neural implicit surfaces beyond ray casting.
Nonetheless, here we reflect on several potential alternatives discussed in \secref{Background}.

Some queries can potentially be implemented by approximating the result with many samples taken randomly or in a regular grid.
The primary disadvantage to this approach is performance; an excessive number of samples may be required, a problem which becomes much worse as resolution increases.
We provide several experimental comparisons to brute-force sampling approaches, showing that our guaranteed range-based queries offer significant benefits (\figref{trace_comparison}, \figref{sampling_perf_plot}, \secref{HierarchicalMeshExtraction}, \secref{IntersectionAndCollisions}, and \secref{ClosestPoint}).

Another possibility is to extract a mesh of the surface, and apply mesh-based techniques.
This too may be an expensive option, and runs the risk of aliasing fine-scale features---typical mesh extraction resolutions are much coarser than the $\delta=0.001$ convergence tolerance used in our experiments.
Nonetheless, when mesh-based computation is used, our fast mesh extraction (\secref{HierarchicalMeshExtraction}) can be used to accelerate the process.

A more principled approach is to leverage a global Lipschitz bound $|\nabla f| < L$.
Intuitively, Lipschitz bounds are a global counterpart to our local range analysis.
However, whereas our range analysis bounds the function locally with respect to each individual evaluation, the Lipschitz constant is computed once for the entire domain, and hence typically gives much less tight bounds.
One popular technique is to estimate $L$ as the product of the maximum eigenvalues of dense layer matrices computed via a power method, which has found applications in deep network regularization and robustness~\cite{arjovsky2017wasserstein,miyato2018spectral,tsuzuku2018lipschitz}.
In our context, replacing $f \to f / L$ would transform any implicit function in to a weak signed distance function.
However, global Lipschitz bounds computed in this manner are extremely pessimistic, on the order of $10^5$ for networks even when they fit high-quality SDFs, making Lipschitz bounds ineffective for guaranteed geometric queries on existing networks.

\section{An Application to Inverse Rendering}
\label{sec:InverseRendering}

Inverse rendering directly optimizes scene data to match target images, encompassing many tasks in computer graphics and vision.
A full review is beyond the scope of this section; we refer to \citet{tewari2021advances}, \citet{li2018differentiable}, and \citet{nicolet2021large}.

Neural implicit surfaces are an appealing representation for surface geometry in inverse rendering, but their use requires somehow intersecting primary rays from a camera against the implicit surface.
Existing work resorts to either raycasting with small, fixed timesteps (\eg{} \cite{niemeyer2020differentiable}), or extracting a mesh and then rasterizing (\eg{} \cite{cole2021differentiable}). 
Our ray casting queries \secref{RayCasting} are a valuable new primitive operation in this context, enabling fast and accurate rendering of a neural implicit surface, even when randomly initialized.

\begin{figure}
\begin{center}
    \includegraphics[width=\columnwidth]{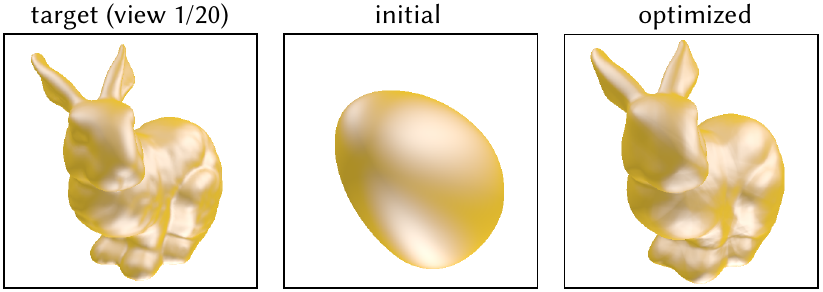}
    \caption{
      Neural implicit surfaces are a promising representation for inverse rendering, but optimization requires casting camera rays in to the surface defined by a not-yet-fitted network---our queries are well-suited to the task.
      In this simple example, a neural implicit shape is optimized to match rendered images of a target surface. 
      The initial view is 100 optimization steps after random initialization.
      Details in \secref{InverseRendering} and \appref{AdditionalDetails}.
      \label{fig:inverse_render_bunny}
    }
\end{center}
\end{figure}

In \figref{inverse_render_bunny} we demonstrate a simple inverse rendering application as a proof of concept.
Here, we fit a neural implicit surface to synthetic target camera views, and render using our ray casting queries and Blinn-Phong shading~\cite{blinn1977models}.
Only two loss terms are used, an $L_1$ image difference loss and a ray occupancy loss which prevents collapse by encouraging the shape to match the foreground mask of the target images (a similar loss appears in \citet[Eqn.\ 14 \& 15]{niemeyer2020differentiable}).
We note that this simple and effective occupancy loss is only available when fitting occupancy networks, as opposed to SDF networks.
Enabling the use of such more general networks is a key goal of this work.
\appref{AdditionalDetails} gives training details.

Though intentionally simple, \figref{inverse_render_bunny} demonstrates the promise of our approach.
In the future, our ray casting primitive could be leveraged in a variety of neural inverse rendering formulations, from deep priors~\cite{mescheder2019occupancy} to global illumination differentiable renderers~\cite{nimier2019mitsuba}. 

\section{Conclusion}

This work studies the application of automatic range analysis to general neural implicit surfaces, enabling a wide variety of useful geometric queries to be applied to existing architectures.

\paragraph{Limitations}

Our queries offer guaranteed accuracy in sense of \secref{Convergence}.
However, we do not address inaccuracies which might arise due to evaluation in inexact floating point arithmetic, although we do not notice any such instabilities in practice.
Existing work in range analysis can bound floating point error via careful manipulation of rounding modes~\cite{stol1997self}, but exact queries would likely demand exact arithmetic, a pursuit which has yielded powerful albeit complex and expensive algorithms elsewhere in computational geometry~\cite{fabri2009cgal}.

There remains a gap between the runtime performance of neural SDF-based queries such as fast ray casting (demonstrated at real-time rates by \citet{takikawa2021neural} and elsewhere), and the more general queries presented here.
Indeed, if one has a high-quality SDF on which sphere tracing is acceptably accurate, it should certainly be used for that purpose.
Rather, the goal of this work is to efficiently extend ray casting---and many other queries---to a much broader set of architectures and applications where a high-quality SDF is unavailable, such as the inverse rendering task in \secref{InverseRendering}.
It is then unsurprising that our more general approach comes at additional computational cost.
In this work we have focused mainly on introducing new algorithms, but further investigation of high-performance kernels could likely accelerate our method significantly.

\paragraph{Future Work}

For now, we primarily study small MLP-like architectures for neural implicit surfaces, but our techniques should also apply more broadly to other architectures with little modification.
Interestingly, architectures which store data in an underlying spatial structure~\cite{takikawa2021neural,mueller2022instant}, could be handled by deriving range bounds for sampling from the structure.
With generalization in mind, we investigated a prototype implementation of range analysis as a fully-automatic transformation which can be applied to arbitrary JAX programs, and we are eager to explore such an approach as a general tool for more exotic architectures.

More broadly, the ability to perform range analysis and build spatial hierarchies likely has value to many other applications of neural fields~\cite{xie2021neural}.
One particular area of interest is neural volumetric rendering~\cite{MildenhallSTBRN20} where we could apply range analysis to hierarchical volumetric integration with guaranteed accuracy.
In general, we hope that providing tools which apply broadly across many architectures and applications will be a key that unlocks many promising research directions.

\begin{acks}

The authors are grateful to Aravind Ramakrishnan, Otman Benche-kroun, and Rohan Sawhney for assistance preparing experiments and reviewing early drafts, as well as Towaki Takikawa, David Levin, and Derek Nowrouzezahrai for insightful discussions.

This research is funded in part by the Fields Institute for Mathematical Sciences, the Vector Institute for AI, the Sloan Research Foundation, NSERC Discovery Grants (RGPIN2017-05235, RGPAS–2017-507938), New Frontiers of Research Fund (NFRFE-201), the Ontario Early Research Award program, the Canada Research Chairs Program, and gifts by Adobe Systems.

\end{acks}

\bibliographystyle{ACM-Reference-Format}
\bibliography{interval-implicits}


\begin{thebibliography}{82}


\ifx \showCODEN    \undefined \def \showCODEN     #1{\unskip}     \fi
\ifx \showDOI      \undefined \def \showDOI       #1{#1}\fi
\ifx \showISBNx    \undefined \def \showISBNx     #1{\unskip}     \fi
\ifx \showISBNxiii \undefined \def \showISBNxiii  #1{\unskip}     \fi
\ifx \showISSN     \undefined \def \showISSN      #1{\unskip}     \fi
\ifx \showLCCN     \undefined \def \showLCCN      #1{\unskip}     \fi
\ifx \shownote     \undefined \def \shownote      #1{#1}          \fi
\ifx \showarticletitle \undefined \def \showarticletitle #1{#1}   \fi
\ifx \showURL      \undefined \def \showURL       {\relax}        \fi
\providecommand\bibfield[2]{#2}
\providecommand\bibinfo[2]{#2}
\providecommand\natexlab[1]{#1}
\providecommand\showeprint[2][]{arXiv:#2}

\bibitem[\protect\citeauthoryear{Adam, Magoulas, Karras, and Vrahatis}{Adam
  et~al\mbox{.}}{2016}]%
        {adam2016bounding}
\bibfield{author}{\bibinfo{person}{Stavros~P Adam}, \bibinfo{person}{George~D
  Magoulas}, \bibinfo{person}{Dimitrios~A Karras}, {and}
  \bibinfo{person}{Michael~N Vrahatis}.} \bibinfo{year}{2016}\natexlab{}.
\newblock \showarticletitle{Bounding the search space for global optimization
  of neural networks learning error: an interval analysis approach}.
\newblock \bibinfo{journal}{\emph{Journal of Machine Learning Research}}
  \bibinfo{volume}{17} (\bibinfo{year}{2016}), \bibinfo{pages}{1--40}.
\newblock


\bibitem[\protect\citeauthoryear{Alefeld and Mayer}{Alefeld and Mayer}{2000}]%
        {alefeld2000interval}
\bibfield{author}{\bibinfo{person}{G{\"o}tz Alefeld} {and}
  \bibinfo{person}{G{\"u}nter Mayer}.} \bibinfo{year}{2000}\natexlab{}.
\newblock \showarticletitle{Interval analysis: theory and applications}.
\newblock \bibinfo{journal}{\emph{Journal of computational and applied
  mathematics}} \bibinfo{volume}{121}, \bibinfo{number}{1-2}
  (\bibinfo{year}{2000}), \bibinfo{pages}{421--464}.
\newblock


\bibitem[\protect\citeauthoryear{Arjovsky, Chintala, and Bottou}{Arjovsky
  et~al\mbox{.}}{2017}]%
        {arjovsky2017wasserstein}
\bibfield{author}{\bibinfo{person}{Martin Arjovsky}, \bibinfo{person}{Soumith
  Chintala}, {and} \bibinfo{person}{L{\'e}on Bottou}.}
  \bibinfo{year}{2017}\natexlab{}.
\newblock \showarticletitle{Wasserstein generative adversarial networks}. In
  \bibinfo{booktitle}{\emph{International conference on machine learning}}.
  PMLR, \bibinfo{pages}{214--223}.
\newblock


\bibitem[\protect\citeauthoryear{Atzmon and Lipman}{Atzmon and Lipman}{2020a}]%
        {atzmon2020sal}
\bibfield{author}{\bibinfo{person}{Matan Atzmon} {and} \bibinfo{person}{Yaron
  Lipman}.} \bibinfo{year}{2020}\natexlab{a}.
\newblock \showarticletitle{Sal: Sign agnostic learning of shapes from raw
  data}. In \bibinfo{booktitle}{\emph{Proceedings of the IEEE/CVF Conference on
  Computer Vision and Pattern Recognition}}. \bibinfo{pages}{2565--2574}.
\newblock


\bibitem[\protect\citeauthoryear{Atzmon and Lipman}{Atzmon and Lipman}{2020b}]%
        {atzmon2020sald}
\bibfield{author}{\bibinfo{person}{Matan Atzmon} {and} \bibinfo{person}{Yaron
  Lipman}.} \bibinfo{year}{2020}\natexlab{b}.
\newblock \showarticletitle{SALD: Sign Agnostic Learning with Derivatives}. In
  \bibinfo{booktitle}{\emph{International Conference on Learning
  Representations}}.
\newblock


\bibitem[\protect\citeauthoryear{B{\'a}lint and Valasek}{B{\'a}lint and
  Valasek}{2018}]%
        {Balint2018Accelerating}
\bibfield{author}{\bibinfo{person}{Csaba B{\'a}lint} {and}
  \bibinfo{person}{G{\'a}bor Valasek}.} \bibinfo{year}{2018}\natexlab{}.
\newblock \showarticletitle{Accelerating Sphere Tracing}.
\newblock \bibinfo{journal}{\emph{Proceedings of Eurographics Short Papers}}
  (\bibinfo{year}{2018}), \bibinfo{pages}{4 pages}.
\newblock
\showISSN{1017-4656}


\bibitem[\protect\citeauthoryear{Barron, Mildenhall, Tancik, Hedman,
  Martin-Brualla, and Srinivasan}{Barron et~al\mbox{.}}{2021}]%
        {barron2021mip}
\bibfield{author}{\bibinfo{person}{Jonathan~T Barron}, \bibinfo{person}{Ben
  Mildenhall}, \bibinfo{person}{Matthew Tancik}, \bibinfo{person}{Peter
  Hedman}, \bibinfo{person}{Ricardo Martin-Brualla}, {and}
  \bibinfo{person}{Pratul~P Srinivasan}.} \bibinfo{year}{2021}\natexlab{}.
\newblock \showarticletitle{Mip-nerf: A multiscale representation for
  anti-aliasing neural radiance fields}. In
  \bibinfo{booktitle}{\emph{Proceedings of the IEEE/CVF International
  Conference on Computer Vision}}. \bibinfo{pages}{5855--5864}.
\newblock


\bibitem[\protect\citeauthoryear{Betancourt and Muhanna}{Betancourt and
  Muhanna}{2021}]%
        {betancourt2021interval}
\bibfield{author}{\bibinfo{person}{David Betancourt} {and}
  \bibinfo{person}{Rafi Muhanna}.} \bibinfo{year}{2021}\natexlab{}.
\newblock \showarticletitle{Interval Deep Learning for Uncertainty
  Quantification in Safety Applications}.
\newblock \bibinfo{journal}{\emph{arXiv preprint arXiv:2105.06438}}
  (\bibinfo{year}{2021}).
\newblock


\bibitem[\protect\citeauthoryear{Blinn}{Blinn}{1977}]%
        {blinn1977models}
\bibfield{author}{\bibinfo{person}{James~F Blinn}.}
  \bibinfo{year}{1977}\natexlab{}.
\newblock \showarticletitle{Models of light reflection for computer synthesized
  pictures}. In \bibinfo{booktitle}{\emph{Proceedings of the 4th annual
  conference on Computer graphics and interactive techniques}}.
  \bibinfo{pages}{192--198}.
\newblock


\bibitem[\protect\citeauthoryear{Blinn}{Blinn}{1982}]%
        {Blinn82}
\bibfield{author}{\bibinfo{person}{James~F. Blinn}.}
  \bibinfo{year}{1982}\natexlab{}.
\newblock \showarticletitle{A Generalization of Algebraic Surface Drawing}.
\newblock \bibinfo{journal}{\emph{{ACM} Trans. Graph.}} \bibinfo{volume}{1},
  \bibinfo{number}{3} (\bibinfo{year}{1982}), \bibinfo{pages}{235--256}.
\newblock


\bibitem[\protect\citeauthoryear{Bloomenthal, Bajaj, Blinn, Cani, Rockwood,
  Wyvill, and Wyvill}{Bloomenthal et~al\mbox{.}}{1997}]%
        {Bloomenthalbook}
\bibfield{author}{\bibinfo{person}{Jules Bloomenthal},
  \bibinfo{person}{Chandrajit Bajaj}, \bibinfo{person}{Jim Blinn},
  \bibinfo{person}{Marie-Paule Cani}, \bibinfo{person}{Alyn Rockwood},
  \bibinfo{person}{Brian Wyvill}, {and} \bibinfo{person}{Geoff Wyvill}.}
  \bibinfo{year}{1997}\natexlab{}.
\newblock \bibinfo{booktitle}{\emph{Introduction to Implicit Surfaces}}.
\newblock


\bibitem[\protect\citeauthoryear{Bradbury, Frostig, Hawkins, Johnson, Leary,
  Maclaurin, Necula, Paszke, Vander{P}las, Wanderman-{M}ilne, and
  Zhang}{Bradbury et~al\mbox{.}}{2018}]%
        {jax2018github}
\bibfield{author}{\bibinfo{person}{James Bradbury}, \bibinfo{person}{Roy
  Frostig}, \bibinfo{person}{Peter Hawkins}, \bibinfo{person}{Matthew~James
  Johnson}, \bibinfo{person}{Chris Leary}, \bibinfo{person}{Dougal Maclaurin},
  \bibinfo{person}{George Necula}, \bibinfo{person}{Adam Paszke},
  \bibinfo{person}{Jake Vander{P}las}, \bibinfo{person}{Skye
  Wanderman-{M}ilne}, {and} \bibinfo{person}{Qiao Zhang}.}
  \bibinfo{year}{2018}\natexlab{}.
\newblock \bibinfo{booktitle}{\emph{{JAX}: composable transformations of
  {P}ython+{N}um{P}y programs}}.
\newblock
\urldef\tempurl%
\url{http://github.com/google/jax}
\showURL{%
\tempurl}


\bibitem[\protect\citeauthoryear{Chan}{Chan}{2008}]%
        {chan2008static}
\bibfield{author}{\bibinfo{person}{Bryan Chan}.}
  \bibinfo{year}{2008}\natexlab{}.
\newblock \emph{\bibinfo{title}{Static Analysis for Efficient Affine Arithmetic
  on GPUs}}.
\newblock \bibinfo{thesistype}{Master's\ thesis}. \bibinfo{school}{University
  of Waterloo}.
\newblock


\bibitem[\protect\citeauthoryear{Chen and Zhang}{Chen and Zhang}{2019}]%
        {ChenZ19}
\bibfield{author}{\bibinfo{person}{Zhiqin Chen} {and} \bibinfo{person}{Hao
  Zhang}.} \bibinfo{year}{2019}\natexlab{}.
\newblock \showarticletitle{Learning Implicit Fields for Generative Shape
  Modeling}. In \bibinfo{booktitle}{\emph{{IEEE} Conference on Computer Vision
  and Pattern Recognition, {CVPR} 2019, Long Beach, CA, USA, June 16-20,
  2019}}. \bibinfo{publisher}{Computer Vision Foundation / {IEEE}},
  \bibinfo{pages}{5939--5948}.
\newblock
\urldef\tempurl%
\url{http://openaccess.thecvf.com/content\_CVPR\_2019/html/Chen\_Learning\_Implicit\_Fields\_for\_Generative\_Shape\_Modeling\_CVPR\_2019\_paper.html}
\showURL{%
\tempurl}


\bibitem[\protect\citeauthoryear{Cole, Genova, Sud, Vlasic, and Zhang}{Cole
  et~al\mbox{.}}{2021}]%
        {cole2021differentiable}
\bibfield{author}{\bibinfo{person}{Forrester Cole}, \bibinfo{person}{Kyle
  Genova}, \bibinfo{person}{Avneesh Sud}, \bibinfo{person}{Daniel Vlasic},
  {and} \bibinfo{person}{Zhoutong Zhang}.} \bibinfo{year}{2021}\natexlab{}.
\newblock \showarticletitle{Differentiable surface rendering via
  non-differentiable sampling}. In \bibinfo{booktitle}{\emph{Proceedings of the
  IEEE/CVF International Conference on Computer Vision}}.
  \bibinfo{pages}{6088--6097}.
\newblock


\bibitem[\protect\citeauthoryear{Comba and Stolfi}{Comba and Stolfi}{1993}]%
        {comba1993affine}
\bibfield{author}{\bibinfo{person}{JLD Comba} {and} \bibinfo{person}{J
  Stolfi}.} \bibinfo{year}{1993}\natexlab{}.
\newblock \bibinfo{title}{Affine arithmetic and its applications to computer
  graphics. Anais do VII SIBGRAPI, 9--18}.
\newblock
\newblock


\bibitem[\protect\citeauthoryear{Dai, Landry, Yang, Pavone, and Tedrake}{Dai
  et~al\mbox{.}}{2021}]%
        {dai2021lyapunov}
\bibfield{author}{\bibinfo{person}{Hongkai Dai}, \bibinfo{person}{Benoit
  Landry}, \bibinfo{person}{Lujie Yang}, \bibinfo{person}{Marco Pavone}, {and}
  \bibinfo{person}{Russ Tedrake}.} \bibinfo{year}{2021}\natexlab{}.
\newblock \showarticletitle{Lyapunov-stable neural-network control}.
\newblock \bibinfo{journal}{\emph{Robotics: Science and Systems}}
  (\bibinfo{year}{2021}).
\newblock


\bibitem[\protect\citeauthoryear{Davies, Nowrouzezahrai, and Jacobson}{Davies
  et~al\mbox{.}}{2020}]%
        {Davies2020}
\bibfield{author}{\bibinfo{person}{Thomas Davies}, \bibinfo{person}{Derek
  Nowrouzezahrai}, {and} \bibinfo{person}{Alec Jacobson}.}
  \bibinfo{year}{2020}\natexlab{}.
\newblock \showarticletitle{On the Effectiveness of Weight-Encoded Neural
  Implicit 3D Shapes}.
\newblock  (\bibinfo{year}{2020}).
\newblock
\urldef\tempurl%
\url{https://arxiv.org/abs/2009.09808}
\showURL{%
\tempurl}


\bibitem[\protect\citeauthoryear{De~Cusatis, De~Figueiredo, and
  Gattass}{De~Cusatis et~al\mbox{.}}{1999}]%
        {de1999interval}
\bibfield{author}{\bibinfo{person}{A De~Cusatis},
  \bibinfo{person}{Luiz~Henrique De~Figueiredo}, {and} \bibinfo{person}{Marcelo
  Gattass}.} \bibinfo{year}{1999}\natexlab{}.
\newblock \showarticletitle{Interval methods for ray casting implicit surfaces
  with affine arithmetic}. In \bibinfo{booktitle}{\emph{XII Brazilian Symposium
  on Computer Graphics and Image Processing (Cat. No. PR00481)}}. IEEE,
  \bibinfo{pages}{65--71}.
\newblock


\bibitem[\protect\citeauthoryear{Duff}{Duff}{1992}]%
        {Duff1992}
\bibfield{author}{\bibinfo{person}{Tom Duff}.} \bibinfo{year}{1992}\natexlab{}.
\newblock \showarticletitle{Interval Arithmetic Recursive Subdivision for
  Implicit Functions and Constructive Solid Geometry}. In
  \bibinfo{booktitle}{\emph{Proceedings of the 19th Annual Conference on
  Computer Graphics and Interactive Techniques}}
  \emph{(\bibinfo{series}{SIGGRAPH '92})}. \bibinfo{publisher}{Association for
  Computing Machinery}, \bibinfo{address}{New York, NY, USA},
  \bibinfo{pages}{131–138}.
\newblock
\showISBNx{0897914791}
\urldef\tempurl%
\url{https://doi.org/10.1145/133994.134027}
\showURL{%
\tempurl}


\bibitem[\protect\citeauthoryear{Elsner, Ibing, Czech, Nehring{-}Wirxel, and
  Kobbelt}{Elsner et~al\mbox{.}}{2021}]%
        {Elsner2021}
\bibfield{author}{\bibinfo{person}{Tim Elsner}, \bibinfo{person}{Moritz Ibing},
  \bibinfo{person}{Victor Czech}, \bibinfo{person}{Julius Nehring{-}Wirxel},
  {and} \bibinfo{person}{Leif Kobbelt}.} \bibinfo{year}{2021}\natexlab{}.
\newblock \showarticletitle{Intuitive Shape Editing in Latent Space}.
\newblock  (\bibinfo{year}{2021}).
\newblock
\urldef\tempurl%
\url{https://arxiv.org/abs/2111.12488}
\showURL{%
\tempurl}


\bibitem[\protect\citeauthoryear{Fabri and Pion}{Fabri and Pion}{2009}]%
        {fabri2009cgal}
\bibfield{author}{\bibinfo{person}{Andreas Fabri} {and}
  \bibinfo{person}{Sylvain Pion}.} \bibinfo{year}{2009}\natexlab{}.
\newblock \showarticletitle{CGAL: The computational geometry algorithms
  library}. In \bibinfo{booktitle}{\emph{Proceedings of the 17th ACM SIGSPATIAL
  international conference on advances in geographic information systems}}.
  \bibinfo{pages}{538--539}.
\newblock


\bibitem[\protect\citeauthoryear{Fl{\'o}rez, Sbert, Sainz, and
  Veh{\'\i}}{Fl{\'o}rez et~al\mbox{.}}{2006}]%
        {florez2006improving}
\bibfield{author}{\bibinfo{person}{Jorge Fl{\'o}rez}, \bibinfo{person}{Mateu
  Sbert}, \bibinfo{person}{Miguel~A Sainz}, {and} \bibinfo{person}{Josep
  Veh{\'\i}}.} \bibinfo{year}{2006}\natexlab{}.
\newblock \showarticletitle{Improving the interval ray tracing of implicit
  surfaces}. In \bibinfo{booktitle}{\emph{Computer Graphics International
  Conference}}. Springer, \bibinfo{pages}{655--664}.
\newblock


\bibitem[\protect\citeauthoryear{Fryazinov, Pasko, and Comninos}{Fryazinov
  et~al\mbox{.}}{2010}]%
        {FryazinovPC10}
\bibfield{author}{\bibinfo{person}{Oleg Fryazinov},
  \bibinfo{person}{Alexander~A. Pasko}, {and} \bibinfo{person}{Peter
  Comninos}.} \bibinfo{year}{2010}\natexlab{}.
\newblock \showarticletitle{Fast reliable interrogation of procedurally defined
  implicit surfaces using extended revised affine arithmetic}.
\newblock \bibinfo{journal}{\emph{Comput. Graph.}} \bibinfo{volume}{34},
  \bibinfo{number}{6} (\bibinfo{year}{2010}), \bibinfo{pages}{708--718}.
\newblock
\urldef\tempurl%
\url{https://doi.org/10.1016/j.cag.2010.07.003}
\showURL{%
\tempurl}


\bibitem[\protect\citeauthoryear{Galin, Gu{\'{e}}rin, Paris, and
  Peytavie}{Galin et~al\mbox{.}}{2020}]%
        {GalinGPP20}
\bibfield{author}{\bibinfo{person}{Eric Galin}, \bibinfo{person}{Eric
  Gu{\'{e}}rin}, \bibinfo{person}{Axel Paris}, {and} \bibinfo{person}{Adrien
  Peytavie}.} \bibinfo{year}{2020}\natexlab{}.
\newblock \showarticletitle{Segment Tracing Using Local Lipschitz Bounds}.
\newblock \bibinfo{journal}{\emph{Comput. Graph. Forum}} \bibinfo{volume}{39},
  \bibinfo{number}{2} (\bibinfo{year}{2020}), \bibinfo{pages}{545--554}.
\newblock
\urldef\tempurl%
\url{https://doi.org/10.1111/cgf.13951}
\showDOI{\tempurl}


\bibitem[\protect\citeauthoryear{Gamito and Maddock}{Gamito and
  Maddock}{2007}]%
        {gamito2007ray}
\bibfield{author}{\bibinfo{person}{Manuel~N Gamito} {and}
  \bibinfo{person}{Steve~C Maddock}.} \bibinfo{year}{2007}\natexlab{}.
\newblock \showarticletitle{Ray casting implicit fractal surfaces with reduced
  affine arithmetic}.
\newblock \bibinfo{journal}{\emph{The Visual Computer}} \bibinfo{volume}{23},
  \bibinfo{number}{3} (\bibinfo{year}{2007}), \bibinfo{pages}{155--165}.
\newblock


\bibitem[\protect\citeauthoryear{Genova, Cole, Sud, Sarna, and
  Funkhouser}{Genova et~al\mbox{.}}{2020}]%
        {GenovaCSSF20}
\bibfield{author}{\bibinfo{person}{Kyle Genova}, \bibinfo{person}{Forrester
  Cole}, \bibinfo{person}{Avneesh Sud}, \bibinfo{person}{Aaron Sarna}, {and}
  \bibinfo{person}{Thomas~A. Funkhouser}.} \bibinfo{year}{2020}\natexlab{}.
\newblock \showarticletitle{Local Deep Implicit Functions for 3D Shape}. In
  \bibinfo{booktitle}{\emph{2020 {IEEE/CVF} Conference on Computer Vision and
  Pattern Recognition, {CVPR} 2020, Seattle, WA, USA, June 13-19, 2020}}.
  \bibinfo{publisher}{Computer Vision Foundation / {IEEE}},
  \bibinfo{pages}{4856--4865}.
\newblock


\bibitem[\protect\citeauthoryear{Gowal, Dvijotham, Stanforth, Bunel, Qin,
  Uesato, Arandjelovic, Mann, and Kohli}{Gowal et~al\mbox{.}}{2018}]%
        {gowal2018effectiveness}
\bibfield{author}{\bibinfo{person}{Sven Gowal}, \bibinfo{person}{Krishnamurthy
  Dvijotham}, \bibinfo{person}{Robert Stanforth}, \bibinfo{person}{Rudy Bunel},
  \bibinfo{person}{Chongli Qin}, \bibinfo{person}{Jonathan Uesato},
  \bibinfo{person}{Relja Arandjelovic}, \bibinfo{person}{Timothy Mann}, {and}
  \bibinfo{person}{Pushmeet Kohli}.} \bibinfo{year}{2018}\natexlab{}.
\newblock \showarticletitle{On the effectiveness of interval bound propagation
  for training verifiably robust models}.
\newblock \bibinfo{journal}{\emph{arXiv preprint arXiv:1810.12715}}
  (\bibinfo{year}{2018}).
\newblock


\bibitem[\protect\citeauthoryear{Gropp, Yariv, Haim, Atzmon, and Lipman}{Gropp
  et~al\mbox{.}}{2020}]%
        {GroppYHAL20}
\bibfield{author}{\bibinfo{person}{Amos Gropp}, \bibinfo{person}{Lior Yariv},
  \bibinfo{person}{Niv Haim}, \bibinfo{person}{Matan Atzmon}, {and}
  \bibinfo{person}{Yaron Lipman}.} \bibinfo{year}{2020}\natexlab{}.
\newblock \showarticletitle{Implicit Geometric Regularization for Learning
  Shapes}. In \bibinfo{booktitle}{\emph{Proceedings of the 37th International
  Conference on Machine Learning, {ICML} 2020, 13-18 July 2020, Virtual Event}}
  \emph{(\bibinfo{series}{Proceedings of Machine Learning Research},
  Vol.~\bibinfo{volume}{119})}. \bibinfo{publisher}{{PMLR}},
  \bibinfo{pages}{3789--3799}.
\newblock


\bibitem[\protect\citeauthoryear{Hart}{Hart}{1996}]%
        {Hart1996Sphere}
\bibfield{author}{\bibinfo{person}{John~C. Hart}.}
  \bibinfo{year}{1996}\natexlab{}.
\newblock \showarticletitle{Sphere Tracing: A Geometric Method for the
  Antialiased Ray Tracing of Implicit Surfaces}.
\newblock \bibinfo{journal}{\emph{The Visual Computer}} \bibinfo{volume}{12},
  \bibinfo{number}{10} (\bibinfo{date}{Dec.} \bibinfo{year}{1996}),
  \bibinfo{pages}{527--545}.
\newblock
\showISSN{1432-2315}
\urldef\tempurl%
\url{https://doi.org/10/b3q2p6}
\showDOI{\tempurl}


\bibitem[\protect\citeauthoryear{Heidrich and Seidel}{Heidrich and
  Seidel}{1998}]%
        {HeidrichS98}
\bibfield{author}{\bibinfo{person}{Wolfgang Heidrich} {and}
  \bibinfo{person}{Hans{-}Peter Seidel}.} \bibinfo{year}{1998}\natexlab{}.
\newblock \showarticletitle{Ray-tracing Procedural Displacement Shaders}. In
  \bibinfo{booktitle}{\emph{Proceedings of the Graphics Interface 1998
  Conference, June 18-20, 1998, Vancouver, BC, Canada}},
  \bibfield{editor}{\bibinfo{person}{Wayne~A. Davis},
  \bibinfo{person}{Kellogg~S. Booth}, {and} \bibinfo{person}{Alain Fournier}}
  (Eds.). \bibinfo{publisher}{Canadian Human-Computer Communications Society},
  \bibinfo{pages}{8--16}.
\newblock


\bibitem[\protect\citeauthoryear{Heidrich, Slusallek, and Seidel}{Heidrich
  et~al\mbox{.}}{1998}]%
        {HeidrichSS98}
\bibfield{author}{\bibinfo{person}{Wolfgang Heidrich}, \bibinfo{person}{Philipp
  Slusallek}, {and} \bibinfo{person}{Hans{-}Peter Seidel}.}
  \bibinfo{year}{1998}\natexlab{}.
\newblock \showarticletitle{Sampling Procedural Shaders Using Affine
  Arithmetic}.
\newblock \bibinfo{journal}{\emph{{ACM} Trans. Graph.}} \bibinfo{volume}{17},
  \bibinfo{number}{3} (\bibinfo{year}{1998}), \bibinfo{pages}{158--176}.
\newblock
\urldef\tempurl%
\url{https://doi.org/10.1145/285857.285859}
\showURL{%
\tempurl}


\bibitem[\protect\citeauthoryear{Jordan and Dimakis}{Jordan and
  Dimakis}{2020}]%
        {JordanD20}
\bibfield{author}{\bibinfo{person}{Matt Jordan} {and}
  \bibinfo{person}{Alexandros~G. Dimakis}.} \bibinfo{year}{2020}\natexlab{}.
\newblock \showarticletitle{Exactly Computing the Local Lipschitz Constant of
  ReLU Networks}. In \bibinfo{booktitle}{\emph{Advances in Neural Information
  Processing Systems 33: Annual Conference on Neural Information Processing
  Systems 2020, NeurIPS 2020, December 6-12, 2020, virtual}},
  \bibfield{editor}{\bibinfo{person}{Hugo Larochelle},
  \bibinfo{person}{Marc'Aurelio Ranzato}, \bibinfo{person}{Raia Hadsell},
  \bibinfo{person}{Maria{-}Florina Balcan}, {and} \bibinfo{person}{Hsuan{-}Tien
  Lin}} (Eds.).
\newblock


\bibitem[\protect\citeauthoryear{Ju, Losasso, Schaefer, and Warren}{Ju
  et~al\mbox{.}}{2002}]%
        {ju2002dual}
\bibfield{author}{\bibinfo{person}{Tao Ju}, \bibinfo{person}{Frank Losasso},
  \bibinfo{person}{Scott Schaefer}, {and} \bibinfo{person}{Joe Warren}.}
  \bibinfo{year}{2002}\natexlab{}.
\newblock \showarticletitle{Dual contouring of hermite data}. In
  \bibinfo{booktitle}{\emph{Proceedings of the 29th annual conference on
  Computer graphics and interactive techniques}}. \bibinfo{pages}{339--346}.
\newblock


\bibitem[\protect\citeauthoryear{Keeter}{Keeter}{2020}]%
        {Keeter2020}
\bibfield{author}{\bibinfo{person}{Matthew~J. Keeter}.}
  \bibinfo{year}{2020}\natexlab{}.
\newblock \showarticletitle{Massively Parallel Rendering of Complex Closed-Form
  Implicit Surfaces}.
\newblock \bibinfo{journal}{\emph{ACM Trans. Graph.}} \bibinfo{volume}{39},
  \bibinfo{number}{4}, Article \bibinfo{articleno}{141} (\bibinfo{date}{jul}
  \bibinfo{year}{2020}), \bibinfo{numpages}{10}~pages.
\newblock
\showISSN{0730-0301}
\urldef\tempurl%
\url{https://doi.org/10.1145/3386569.3392429}
\showURL{%
\tempurl}


\bibitem[\protect\citeauthoryear{Keinert, Sch{\"a}fer, Kornd{\"o}rfer, Ganse,
  and Stamminger}{Keinert et~al\mbox{.}}{2013}]%
        {Keinert2013Improved}
\bibfield{author}{\bibinfo{person}{Benjamin Keinert}, \bibinfo{person}{Henry
  Sch{\"a}fer}, \bibinfo{person}{Johann Kornd{\"o}rfer}, \bibinfo{person}{Urs
  Ganse}, {and} \bibinfo{person}{Marc Stamminger}.}
  \bibinfo{year}{2013}\natexlab{}.
\newblock \showarticletitle{Improved Ray Casting of Procedural Distance
  Bounds}.
\newblock \bibinfo{journal}{\emph{Journal of Graphics Tools}}
  \bibinfo{volume}{17}, \bibinfo{number}{4} (\bibinfo{date}{Oct.}
  \bibinfo{year}{2013}), \bibinfo{pages}{127--138}.
\newblock
\showISSN{2165-347X}


\bibitem[\protect\citeauthoryear{Knoll, Hijazi, Hansen, Wald, and Hagen}{Knoll
  et~al\mbox{.}}{2007}]%
        {Knoll2007}
\bibfield{author}{\bibinfo{person}{Aaron Knoll}, \bibinfo{person}{Younis
  Hijazi}, \bibinfo{person}{Charles Hansen}, \bibinfo{person}{Ingo Wald}, {and}
  \bibinfo{person}{Hans Hagen}.} \bibinfo{year}{2007}\natexlab{}.
\newblock \showarticletitle{Interactive Ray Tracing of Arbitrary Implicits with
  SIMD Interval Arithmetic}. In \bibinfo{booktitle}{\emph{2007 IEEE Symposium
  on Interactive Ray Tracing}}. \bibinfo{pages}{11--18}.
\newblock


\bibitem[\protect\citeauthoryear{Knoll, Hijazi, Kensler, Schott, Hansen, and
  Hagen}{Knoll et~al\mbox{.}}{2009}]%
        {KnollHKSHH09}
\bibfield{author}{\bibinfo{person}{Aaron Knoll}, \bibinfo{person}{Younis
  Hijazi}, \bibinfo{person}{Andrew~E. Kensler}, \bibinfo{person}{Mathias
  Schott}, \bibinfo{person}{Charles~D. Hansen}, {and} \bibinfo{person}{Hans
  Hagen}.} \bibinfo{year}{2009}\natexlab{}.
\newblock \showarticletitle{Fast Ray Tracing of Arbitrary Implicit Surfaces
  with Interval and Affine Arithmetic}.
\newblock \bibinfo{journal}{\emph{Comput. Graph. Forum}} \bibinfo{volume}{28},
  \bibinfo{number}{1} (\bibinfo{year}{2009}), \bibinfo{pages}{26--40}.
\newblock
\urldef\tempurl%
\url{https://doi.org/10.1111/j.1467-8659.2008.01189.x}
\showDOI{\tempurl}


\bibitem[\protect\citeauthoryear{Lei and Jia}{Lei and Jia}{2020}]%
        {LeiJ20}
\bibfield{author}{\bibinfo{person}{Jiabao Lei} {and} \bibinfo{person}{Kui
  Jia}.} \bibinfo{year}{2020}\natexlab{}.
\newblock \showarticletitle{Analytic Marching: An Analytic Meshing Solution
  from Deep Implicit Surface Networks}. In
  \bibinfo{booktitle}{\emph{Proceedings of the 37th International Conference on
  Machine Learning, {ICML} 2020, 13-18 July 2020, Virtual Event}}
  \emph{(\bibinfo{series}{Proceedings of Machine Learning Research},
  Vol.~\bibinfo{volume}{119})}. \bibinfo{publisher}{{PMLR}},
  \bibinfo{pages}{5789--5798}.
\newblock


\bibitem[\protect\citeauthoryear{Li, Aittala, Durand, and Lehtinen}{Li
  et~al\mbox{.}}{2018}]%
        {li2018differentiable}
\bibfield{author}{\bibinfo{person}{Tzu-Mao Li}, \bibinfo{person}{Miika
  Aittala}, \bibinfo{person}{Fr{\'e}do Durand}, {and} \bibinfo{person}{Jaakko
  Lehtinen}.} \bibinfo{year}{2018}\natexlab{}.
\newblock \showarticletitle{Differentiable monte carlo ray tracing through edge
  sampling}.
\newblock \bibinfo{journal}{\emph{ACM Transactions on Graphics (TOG)}}
  \bibinfo{volume}{37}, \bibinfo{number}{6} (\bibinfo{year}{2018}),
  \bibinfo{pages}{1--11}.
\newblock


\bibitem[\protect\citeauthoryear{Liao, Donne, and Geiger}{Liao
  et~al\mbox{.}}{2018}]%
        {liao2018deep}
\bibfield{author}{\bibinfo{person}{Yiyi Liao}, \bibinfo{person}{Simon Donne},
  {and} \bibinfo{person}{Andreas Geiger}.} \bibinfo{year}{2018}\natexlab{}.
\newblock \showarticletitle{Deep marching cubes: Learning explicit surface
  representations}. In \bibinfo{booktitle}{\emph{Proceedings of the IEEE
  Conference on Computer Vision and Pattern Recognition}}.
  \bibinfo{pages}{2916--2925}.
\newblock


\bibitem[\protect\citeauthoryear{Lorensen and Cline}{Lorensen and
  Cline}{1987}]%
        {lorensen1987marching}
\bibfield{author}{\bibinfo{person}{William~E Lorensen} {and}
  \bibinfo{person}{Harvey~E Cline}.} \bibinfo{year}{1987}\natexlab{}.
\newblock \showarticletitle{Marching cubes: A high resolution 3D surface
  construction algorithm}.
\newblock \bibinfo{journal}{\emph{ACM siggraph computer graphics}}
  \bibinfo{volume}{21}, \bibinfo{number}{4} (\bibinfo{year}{1987}),
  \bibinfo{pages}{163--169}.
\newblock


\bibitem[\protect\citeauthoryear{Menon}{Menon}{1996}]%
        {Menonbook}
\bibfield{author}{\bibinfo{person}{Jai Menon}.}
  \bibinfo{year}{1996}\natexlab{}.
\newblock \bibinfo{booktitle}{\emph{An Introduction to Implicit Techniques}}.
\newblock


\bibitem[\protect\citeauthoryear{Mescheder, Oechsle, Niemeyer, Nowozin, and
  Geiger}{Mescheder et~al\mbox{.}}{2019}]%
        {mescheder2019occupancy}
\bibfield{author}{\bibinfo{person}{Lars Mescheder}, \bibinfo{person}{Michael
  Oechsle}, \bibinfo{person}{Michael Niemeyer}, \bibinfo{person}{Sebastian
  Nowozin}, {and} \bibinfo{person}{Andreas Geiger}.}
  \bibinfo{year}{2019}\natexlab{}.
\newblock \showarticletitle{Occupancy networks: Learning 3d reconstruction in
  function space}. In \bibinfo{booktitle}{\emph{Proceedings of the IEEE/CVF
  Conference on Computer Vision and Pattern Recognition}}.
  \bibinfo{pages}{4460--4470}.
\newblock


\bibitem[\protect\citeauthoryear{Mildenhall, Srinivasan, Tancik, Barron,
  Ramamoorthi, and Ng}{Mildenhall et~al\mbox{.}}{2020}]%
        {MildenhallSTBRN20}
\bibfield{author}{\bibinfo{person}{Ben Mildenhall}, \bibinfo{person}{Pratul~P.
  Srinivasan}, \bibinfo{person}{Matthew Tancik}, \bibinfo{person}{Jonathan~T.
  Barron}, \bibinfo{person}{Ravi Ramamoorthi}, {and} \bibinfo{person}{Ren Ng}.}
  \bibinfo{year}{2020}\natexlab{}.
\newblock \showarticletitle{NeRF: Representing Scenes as Neural Radiance Fields
  for View Synthesis}. In \bibinfo{booktitle}{\emph{{ECCV} 2020}},
  Vol.~\bibinfo{volume}{12346}. \bibinfo{publisher}{Springer},
  \bibinfo{pages}{405--421}.
\newblock


\bibitem[\protect\citeauthoryear{Mirman, Baader, and Vechev}{Mirman
  et~al\mbox{.}}{2021}]%
        {mirman2021fundamental}
\bibfield{author}{\bibinfo{person}{Matthew Mirman}, \bibinfo{person}{Maximilian
  Baader}, {and} \bibinfo{person}{Martin Vechev}.}
  \bibinfo{year}{2021}\natexlab{}.
\newblock \showarticletitle{The Fundamental Limits of Interval Arithmetic for
  Neural Networks}.
\newblock \bibinfo{journal}{\emph{arXiv preprint arXiv:2112.05235}}
  (\bibinfo{year}{2021}).
\newblock


\bibitem[\protect\citeauthoryear{Mitchell}{Mitchell}{1990}]%
        {mitchell1990robust}
\bibfield{author}{\bibinfo{person}{Don~P Mitchell}.}
  \bibinfo{year}{1990}\natexlab{}.
\newblock \showarticletitle{Robust ray intersection with interval arithmetic}.
  In \bibinfo{booktitle}{\emph{Proceedings of Graphics Interface}},
  Vol.~\bibinfo{volume}{90}. \bibinfo{pages}{68--74}.
\newblock


\bibitem[\protect\citeauthoryear{Miyato, Kataoka, Koyama, and Yoshida}{Miyato
  et~al\mbox{.}}{2018}]%
        {miyato2018spectral}
\bibfield{author}{\bibinfo{person}{Takeru Miyato}, \bibinfo{person}{Toshiki
  Kataoka}, \bibinfo{person}{Masanori Koyama}, {and} \bibinfo{person}{Yuichi
  Yoshida}.} \bibinfo{year}{2018}\natexlab{}.
\newblock \showarticletitle{Spectral normalization for generative adversarial
  networks}.
\newblock \bibinfo{journal}{\emph{arXiv preprint arXiv:1802.05957}}
  (\bibinfo{year}{2018}).
\newblock


\bibitem[\protect\citeauthoryear{Moore, Kearfott, and Cloud}{Moore
  et~al\mbox{.}}{2009}]%
        {moore2009introduction}
\bibfield{author}{\bibinfo{person}{Ramon~E Moore}, \bibinfo{person}{R~Baker
  Kearfott}, {and} \bibinfo{person}{Michael~J Cloud}.}
  \bibinfo{year}{2009}\natexlab{}.
\newblock \bibinfo{booktitle}{\emph{Introduction to interval analysis}}.
\newblock \bibinfo{publisher}{SIAM}.
\newblock


\bibitem[\protect\citeauthoryear{M\"uller, Evans, Schied, and Keller}{M\"uller
  et~al\mbox{.}}{2022}]%
        {mueller2022instant}
\bibfield{author}{\bibinfo{person}{Thomas M\"uller}, \bibinfo{person}{Alex
  Evans}, \bibinfo{person}{Christoph Schied}, {and} \bibinfo{person}{Alexander
  Keller}.} \bibinfo{year}{2022}\natexlab{}.
\newblock \showarticletitle{Instant Neural Graphics Primitives with a
  Multiresolution Hash Encoding}.
\newblock \bibinfo{journal}{\emph{arXiv:2201.05989}} (\bibinfo{date}{Jan.}
  \bibinfo{year}{2022}).
\newblock


\bibitem[\protect\citeauthoryear{Nabizadeh, Ramamoorthi, and Chern}{Nabizadeh
  et~al\mbox{.}}{2021}]%
        {nabizadeh2021kelvin}
\bibfield{author}{\bibinfo{person}{Mohammad~Sina Nabizadeh},
  \bibinfo{person}{Ravi Ramamoorthi}, {and} \bibinfo{person}{Albert Chern}.}
  \bibinfo{year}{2021}\natexlab{}.
\newblock \showarticletitle{Kelvin transformations for simulations on infinite
  domains}.
\newblock \bibinfo{journal}{\emph{ACM Transactions on Graphics (TOG)}}
  \bibinfo{volume}{40}, \bibinfo{number}{4} (\bibinfo{year}{2021}),
  \bibinfo{pages}{1--15}.
\newblock


\bibitem[\protect\citeauthoryear{Nicolet, Jacobson, and Jakob}{Nicolet
  et~al\mbox{.}}{2021}]%
        {nicolet2021large}
\bibfield{author}{\bibinfo{person}{Baptiste Nicolet}, \bibinfo{person}{Alec
  Jacobson}, {and} \bibinfo{person}{Wenzel Jakob}.}
  \bibinfo{year}{2021}\natexlab{}.
\newblock \showarticletitle{Large steps in inverse rendering of geometry}.
\newblock \bibinfo{journal}{\emph{ACM Transactions on Graphics (TOG)}}
  \bibinfo{volume}{40}, \bibinfo{number}{6} (\bibinfo{year}{2021}),
  \bibinfo{pages}{1--13}.
\newblock


\bibitem[\protect\citeauthoryear{Niemeyer, Mescheder, Oechsle, and
  Geiger}{Niemeyer et~al\mbox{.}}{2020}]%
        {niemeyer2020differentiable}
\bibfield{author}{\bibinfo{person}{Michael Niemeyer}, \bibinfo{person}{Lars
  Mescheder}, \bibinfo{person}{Michael Oechsle}, {and} \bibinfo{person}{Andreas
  Geiger}.} \bibinfo{year}{2020}\natexlab{}.
\newblock \showarticletitle{Differentiable volumetric rendering: Learning
  implicit 3d representations without 3d supervision}. In
  \bibinfo{booktitle}{\emph{Proceedings of the IEEE/CVF Conference on Computer
  Vision and Pattern Recognition}}. \bibinfo{pages}{3504--3515}.
\newblock


\bibitem[\protect\citeauthoryear{Nimier-David, Vicini, Zeltner, and
  Jakob}{Nimier-David et~al\mbox{.}}{2019}]%
        {nimier2019mitsuba}
\bibfield{author}{\bibinfo{person}{Merlin Nimier-David}, \bibinfo{person}{Delio
  Vicini}, \bibinfo{person}{Tizian Zeltner}, {and} \bibinfo{person}{Wenzel
  Jakob}.} \bibinfo{year}{2019}\natexlab{}.
\newblock \showarticletitle{Mitsuba 2: A retargetable forward and inverse
  renderer}.
\newblock \bibinfo{journal}{\emph{ACM Transactions on Graphics (TOG)}}
  \bibinfo{volume}{38}, \bibinfo{number}{6} (\bibinfo{year}{2019}),
  \bibinfo{pages}{1--17}.
\newblock


\bibitem[\protect\citeauthoryear{Park, Florence, Straub, Newcombe, and
  Lovegrove}{Park et~al\mbox{.}}{2019}]%
        {park2019deepsdf}
\bibfield{author}{\bibinfo{person}{Jeong~Joon Park}, \bibinfo{person}{Peter
  Florence}, \bibinfo{person}{Julian Straub}, \bibinfo{person}{Richard
  Newcombe}, {and} \bibinfo{person}{Steven Lovegrove}.}
  \bibinfo{year}{2019}\natexlab{}.
\newblock \showarticletitle{Deepsdf: Learning continuous signed distance
  functions for shape representation}. In \bibinfo{booktitle}{\emph{Proceedings
  of the IEEE/CVF Conference on Computer Vision and Pattern Recognition}}.
  \bibinfo{pages}{165--174}.
\newblock


\bibitem[\protect\citeauthoryear{Perlin and Hoffert}{Perlin and
  Hoffert}{1989}]%
        {PerlinH89}
\bibfield{author}{\bibinfo{person}{Ken Perlin} {and} \bibinfo{person}{Eric~M.
  Hoffert}.} \bibinfo{year}{1989}\natexlab{}.
\newblock \showarticletitle{Hypertexture}. In
  \bibinfo{booktitle}{\emph{Proceedings of the 16th Annual Conference on
  Computer Graphics and Interactive Techniques, {SIGGRAPH} 1989, Boston, MA,
  USA, July 31 - August 4, 1989}}, \bibfield{editor}{\bibinfo{person}{James~J.
  Thomas}} (Ed.). \bibinfo{publisher}{{ACM}}, \bibinfo{pages}{253--262}.
\newblock


\bibitem[\protect\citeauthoryear{Proszewska, Mazur, Trzci{\'n}ski, and
  Spurek}{Proszewska et~al\mbox{.}}{2021}]%
        {proszewska2021hypercube}
\bibfield{author}{\bibinfo{person}{Magdalena Proszewska},
  \bibinfo{person}{Marcin Mazur}, \bibinfo{person}{Tomasz Trzci{\'n}ski}, {and}
  \bibinfo{person}{Przemys{\l}aw Spurek}.} \bibinfo{year}{2021}\natexlab{}.
\newblock \showarticletitle{HyperCube: Implicit Field Representations of
  Voxelized 3D Models}.
\newblock \bibinfo{journal}{\emph{arXiv preprint arXiv:2110.05770}}
  (\bibinfo{year}{2021}).
\newblock


\bibitem[\protect\citeauthoryear{Quilez}{Quilez}{2008}]%
        {Inigo2008}
\bibfield{author}{\bibinfo{person}{Inigo Quilez}.}
  \bibinfo{year}{2008}\natexlab{}.
\newblock \bibinfo{title}{3D SDF functions}.
\newblock
\newblock
\urldef\tempurl%
\url{https://www.iquilezles.org/www/articles/distfunctions/distfunctions.htm}
\showURL{%
\tempurl}


\bibitem[\protect\citeauthoryear{Ratz}{Ratz}{1996}]%
        {ratz1996optimized}
\bibfield{author}{\bibinfo{person}{Dietmar Ratz}.}
  \bibinfo{year}{1996}\natexlab{}.
\newblock \bibinfo{booktitle}{\emph{An optimized interval slope arithmetic and
  its application}}.
\newblock \bibinfo{publisher}{Inst. f{\"u}r Angewandte Mathematik}.
\newblock


\bibitem[\protect\citeauthoryear{Reiner, M{\"u}ckl, and Dachsbacher}{Reiner
  et~al\mbox{.}}{2011}]%
        {Reiner2011Interactive}
\bibfield{author}{\bibinfo{person}{Tim Reiner}, \bibinfo{person}{Gregor
  M{\"u}ckl}, {and} \bibinfo{person}{Carsten Dachsbacher}.}
  \bibinfo{year}{2011}\natexlab{}.
\newblock \showarticletitle{Interactive Modeling of Implicit Surfaces Using a
  Direct Visualization Approach with Signed Distance Functions}.
\newblock \bibinfo{journal}{\emph{Computers and Graphics}}
  \bibinfo{volume}{35}, \bibinfo{number}{3} (\bibinfo{date}{June}
  \bibinfo{year}{2011}), \bibinfo{pages}{596--603}.
\newblock
\showISSN{0097-8493}


\bibitem[\protect\citeauthoryear{Remelli, Lukoianov, Richter, Guillard,
  Bagautdinov, Baqu{\'{e}}, and Fua}{Remelli et~al\mbox{.}}{2020}]%
        {RemelliLRGBBF20}
\bibfield{author}{\bibinfo{person}{Edoardo Remelli}, \bibinfo{person}{Artem
  Lukoianov}, \bibinfo{person}{Stephan~R. Richter},
  \bibinfo{person}{Beno{\^{\i}}t Guillard}, \bibinfo{person}{Timur~M.
  Bagautdinov}, \bibinfo{person}{Pierre Baqu{\'{e}}}, {and}
  \bibinfo{person}{Pascal Fua}.} \bibinfo{year}{2020}\natexlab{}.
\newblock \showarticletitle{MeshSDF: Differentiable Iso-Surface Extraction}. In
  \bibinfo{booktitle}{\emph{Advances in Neural Information Processing Systems
  33: Annual Conference on Neural Information Processing Systems 2020, NeurIPS
  2020, December 6-12, 2020, virtual}}.
\newblock


\bibitem[\protect\citeauthoryear{Reshetov, Soupikov, and Hurley}{Reshetov
  et~al\mbox{.}}{2005}]%
        {reshetov2005multi}
\bibfield{author}{\bibinfo{person}{Alexander Reshetov}, \bibinfo{person}{Alexei
  Soupikov}, {and} \bibinfo{person}{Jim Hurley}.}
  \bibinfo{year}{2005}\natexlab{}.
\newblock \showarticletitle{Multi-level ray tracing algorithm}.
\newblock \bibinfo{journal}{\emph{ACM Transactions on Graphics (TOG)}}
  \bibinfo{volume}{24}, \bibinfo{number}{3} (\bibinfo{year}{2005}),
  \bibinfo{pages}{1176--1185}.
\newblock


\bibitem[\protect\citeauthoryear{Ricci}{Ricci}{1973}]%
        {Ricci73}
\bibfield{author}{\bibinfo{person}{A. Ricci}.} \bibinfo{year}{1973}\natexlab{}.
\newblock \showarticletitle{A Constructive Geometry for Computer Graphics}.
\newblock \bibinfo{journal}{\emph{Comput. J.}} \bibinfo{volume}{16},
  \bibinfo{number}{2} (\bibinfo{year}{1973}), \bibinfo{pages}{157--160}.
\newblock
\urldef\tempurl%
\url{https://doi.org/10.1093/comjnl/16.2.157}
\showDOI{\tempurl}


\bibitem[\protect\citeauthoryear{Rump}{Rump}{1999}]%
        {rump1999intlab}
\bibfield{author}{\bibinfo{person}{Siegfried~M Rump}.}
  \bibinfo{year}{1999}\natexlab{}.
\newblock \showarticletitle{INTLAB—interval laboratory}.
\newblock In \bibinfo{booktitle}{\emph{Developments in reliable computing}}.
  \bibinfo{publisher}{Springer}, \bibinfo{pages}{77--104}.
\newblock


\bibitem[\protect\citeauthoryear{Rump and Kashiwagi}{Rump and
  Kashiwagi}{2015}]%
        {rump2015implementation}
\bibfield{author}{\bibinfo{person}{Siegfried~M Rump} {and}
  \bibinfo{person}{Masahide Kashiwagi}.} \bibinfo{year}{2015}\natexlab{}.
\newblock \showarticletitle{Implementation and improvements of affine
  arithmetic}.
\newblock \bibinfo{journal}{\emph{Nonlinear Theory and Its Applications,
  IEICE}} \bibinfo{volume}{6}, \bibinfo{number}{3} (\bibinfo{year}{2015}),
  \bibinfo{pages}{341--359}.
\newblock


\bibitem[\protect\citeauthoryear{Sahoo, Das, and Chakraverty}{Sahoo
  et~al\mbox{.}}{2015}]%
        {sahoo2015interval}
\bibfield{author}{\bibinfo{person}{Deepti~Moyi Sahoo},
  \bibinfo{person}{Abhishek Das}, {and} \bibinfo{person}{Snehashish
  Chakraverty}.} \bibinfo{year}{2015}\natexlab{}.
\newblock \showarticletitle{Interval data-based system identification of
  multistorey shear buildings by artificial neural network modelling}.
\newblock \bibinfo{journal}{\emph{Architectural Science Review}}
  \bibinfo{volume}{58}, \bibinfo{number}{3} (\bibinfo{year}{2015}),
  \bibinfo{pages}{244--254}.
\newblock


\bibitem[\protect\citeauthoryear{Sawhney and Crane}{Sawhney and Crane}{2020}]%
        {Sawhney:2020:MCG}
\bibfield{author}{\bibinfo{person}{Rohan Sawhney} {and} \bibinfo{person}{Keenan
  Crane}.} \bibinfo{year}{2020}\natexlab{}.
\newblock \showarticletitle{Monte Carlo Geometry Processing: A Grid-Free
  Approach to PDE-Based Methods on Volumetric Domains}.
\newblock \bibinfo{journal}{\emph{ACM Trans. Graph.}} \bibinfo{volume}{39},
  \bibinfo{number}{4} (\bibinfo{year}{2020}).
\newblock


\bibitem[\protect\citeauthoryear{Seyb, Jacobson, Nowrouzezahrai, and
  Jarosz}{Seyb et~al\mbox{.}}{2019}]%
        {SeybJNJ19}
\bibfield{author}{\bibinfo{person}{Dario Seyb}, \bibinfo{person}{Alec
  Jacobson}, \bibinfo{person}{Derek Nowrouzezahrai}, {and}
  \bibinfo{person}{Wojciech Jarosz}.} \bibinfo{year}{2019}\natexlab{}.
\newblock \showarticletitle{Non-linear sphere tracing for rendering deformed
  signed distance fields}.
\newblock \bibinfo{journal}{\emph{{ACM} Trans. Graph.}} \bibinfo{volume}{38},
  \bibinfo{number}{6} (\bibinfo{year}{2019}), \bibinfo{pages}{229:1--229:12}.
\newblock


\bibitem[\protect\citeauthoryear{Shen, Gao, Yin, Liu, and Fidler}{Shen
  et~al\mbox{.}}{2021}]%
        {shen2021deep}
\bibfield{author}{\bibinfo{person}{Tianchang Shen}, \bibinfo{person}{Jun Gao},
  \bibinfo{person}{Kangxue Yin}, \bibinfo{person}{Ming-Yu Liu}, {and}
  \bibinfo{person}{Sanja Fidler}.} \bibinfo{year}{2021}\natexlab{}.
\newblock \showarticletitle{Deep Marching Tetrahedra: a Hybrid Representation
  for High-Resolution 3D Shape Synthesis}.
\newblock \bibinfo{journal}{\emph{Advances in Neural Information Processing
  Systems}}  \bibinfo{volume}{34} (\bibinfo{year}{2021}).
\newblock


\bibitem[\protect\citeauthoryear{Stolfi and De~Figueiredo}{Stolfi and
  De~Figueiredo}{1997}]%
        {stol1997self}
\bibfield{author}{\bibinfo{person}{Jorge Stolfi} {and}
  \bibinfo{person}{Luiz~Henrique De~Figueiredo}.}
  \bibinfo{year}{1997}\natexlab{}.
\newblock \showarticletitle{Self-validated numerical methods and applications}.
  In \bibinfo{booktitle}{\emph{Monograph for 21st Brazilian Mathematics
  Colloquium, IMPA, Rio de Janeiro. Citeseer}}, Vol.~\bibinfo{volume}{5}.
  Citeseer.
\newblock


\bibitem[\protect\citeauthoryear{Takikawa, Litalien, Yin, Kreis, Loop,
  Nowrouzezahrai, Jacobson, McGuire, and Fidler}{Takikawa
  et~al\mbox{.}}{2021}]%
        {takikawa2021neural}
\bibfield{author}{\bibinfo{person}{Towaki Takikawa}, \bibinfo{person}{Joey
  Litalien}, \bibinfo{person}{Kangxue Yin}, \bibinfo{person}{Karsten Kreis},
  \bibinfo{person}{Charles Loop}, \bibinfo{person}{Derek Nowrouzezahrai},
  \bibinfo{person}{Alec Jacobson}, \bibinfo{person}{Morgan McGuire}, {and}
  \bibinfo{person}{Sanja Fidler}.} \bibinfo{year}{2021}\natexlab{}.
\newblock \showarticletitle{Neural geometric level of detail: Real-time
  rendering with implicit 3D shapes}. In \bibinfo{booktitle}{\emph{Proceedings
  of the IEEE/CVF Conference on Computer Vision and Pattern Recognition}}.
  \bibinfo{pages}{11358--11367}.
\newblock


\bibitem[\protect\citeauthoryear{Tewari, Thies, Mildenhall, Srinivasan,
  Tretschk, Wang, Lassner, Sitzmann, Martin-Brualla, Lombardi,
  et~al\mbox{.}}{Tewari et~al\mbox{.}}{2021}]%
        {tewari2021advances}
\bibfield{author}{\bibinfo{person}{Ayush Tewari}, \bibinfo{person}{Justus
  Thies}, \bibinfo{person}{Ben Mildenhall}, \bibinfo{person}{Pratul
  Srinivasan}, \bibinfo{person}{Edgar Tretschk}, \bibinfo{person}{Yifan Wang},
  \bibinfo{person}{Christoph Lassner}, \bibinfo{person}{Vincent Sitzmann},
  \bibinfo{person}{Ricardo Martin-Brualla}, \bibinfo{person}{Stephen Lombardi},
  {et~al\mbox{.}}} \bibinfo{year}{2021}\natexlab{}.
\newblock \showarticletitle{Advances in neural rendering}.
\newblock \bibinfo{journal}{\emph{arXiv preprint arXiv:2111.05849}}
  (\bibinfo{year}{2021}).
\newblock


\bibitem[\protect\citeauthoryear{Thonat, Beaune, Sun, Carr, and
  Boubekeur}{Thonat et~al\mbox{.}}{2021}]%
        {TBSCB:2021:TFDM}
\bibfield{author}{\bibinfo{person}{Theo Thonat}, \bibinfo{person}{Francois
  Beaune}, \bibinfo{person}{Xin Sun}, \bibinfo{person}{Nathan Carr}, {and}
  \bibinfo{person}{Tamy Boubekeur}.} \bibinfo{year}{2021}\natexlab{}.
\newblock \showarticletitle{Tessellation-Free Displacement Mapping for Ray
  Tracing}.
\newblock  \bibinfo{volume}{40}, \bibinfo{number}{6}, Article
  \bibinfo{articleno}{282} (\bibinfo{date}{dec} \bibinfo{year}{2021}),
  \bibinfo{numpages}{16}~pages.
\newblock
\showISSN{0730-0301}
\urldef\tempurl%
\url{https://doi.org/10.1145/3478513.3480535}
\showDOI{\tempurl}


\bibitem[\protect\citeauthoryear{Tsuzuku, Sato, and Sugiyama}{Tsuzuku
  et~al\mbox{.}}{2018}]%
        {tsuzuku2018lipschitz}
\bibfield{author}{\bibinfo{person}{Yusuke Tsuzuku}, \bibinfo{person}{Issei
  Sato}, {and} \bibinfo{person}{Masashi Sugiyama}.}
  \bibinfo{year}{2018}\natexlab{}.
\newblock \showarticletitle{Lipschitz-margin training: scalable certification
  of perturbation invariance for deep neural networks}. In
  \bibinfo{booktitle}{\emph{Proceedings of the 32nd International Conference on
  Neural Information Processing Systems}}. \bibinfo{pages}{6542--6551}.
\newblock


\bibitem[\protect\citeauthoryear{Virmaux and Scaman}{Virmaux and
  Scaman}{2018}]%
        {VirmauxS18}
\bibfield{author}{\bibinfo{person}{Aladin Virmaux} {and} \bibinfo{person}{Kevin
  Scaman}.} \bibinfo{year}{2018}\natexlab{}.
\newblock \showarticletitle{Lipschitz regularity of deep neural networks:
  analysis and efficient estimation}. In \bibinfo{booktitle}{\emph{Advances in
  Neural Information Processing Systems 31: Annual Conference on Neural
  Information Processing Systems 2018, NeurIPS 2018, December 3-8, 2018,
  Montr{\'{e}}al, Canada}}. \bibinfo{pages}{3839--3848}.
\newblock


\bibitem[\protect\citeauthoryear{Wang, Wu, {\"{O}}ztireli, and
  Sorkine{-}Hornung}{Wang et~al\mbox{.}}{2021}]%
        {WangWOS21}
\bibfield{author}{\bibinfo{person}{Yifan Wang}, \bibinfo{person}{Shihao Wu},
  \bibinfo{person}{Cengiz {\"{O}}ztireli}, {and} \bibinfo{person}{Olga
  Sorkine{-}Hornung}.} \bibinfo{year}{2021}\natexlab{}.
\newblock \showarticletitle{Iso-Points: Optimizing Neural Implicit Surfaces
  With Hybrid Representations}. In \bibinfo{booktitle}{\emph{{IEEE} Conference
  on Computer Vision and Pattern Recognition, {CVPR} 2021, virtual, June 19-25,
  2021}}. \bibinfo{publisher}{Computer Vision Foundation / {IEEE}},
  \bibinfo{pages}{374--383}.
\newblock


\bibitem[\protect\citeauthoryear{Wyvill, McPheeters, and Wyvill}{Wyvill
  et~al\mbox{.}}{1986}]%
        {WyvillMW86}
\bibfield{author}{\bibinfo{person}{Geoff Wyvill}, \bibinfo{person}{Craig
  McPheeters}, {and} \bibinfo{person}{Brian Wyvill}.}
  \bibinfo{year}{1986}\natexlab{}.
\newblock \showarticletitle{Data structure for \emph{soft} objects}.
\newblock \bibinfo{journal}{\emph{Vis. Comput.}} \bibinfo{volume}{2},
  \bibinfo{number}{4} (\bibinfo{year}{1986}), \bibinfo{pages}{227--234}.
\newblock
\urldef\tempurl%
\url{https://doi.org/10.1007/BF01900346}
\showDOI{\tempurl}


\bibitem[\protect\citeauthoryear{Xie, Takikawa, Saito, Litany, Yan, Khan,
  Tombari, Tompkin, Sitzmann, and Sridhar}{Xie et~al\mbox{.}}{2022}]%
        {xie2021neural}
\bibfield{author}{\bibinfo{person}{Yiheng Xie}, \bibinfo{person}{Towaki
  Takikawa}, \bibinfo{person}{Shunsuke Saito}, \bibinfo{person}{Or Litany},
  \bibinfo{person}{Shiqin Yan}, \bibinfo{person}{Numair Khan},
  \bibinfo{person}{Federico Tombari}, \bibinfo{person}{James Tompkin},
  \bibinfo{person}{Vincent Sitzmann}, {and} \bibinfo{person}{Srinath Sridhar}.}
  \bibinfo{year}{2022}\natexlab{}.
\newblock \showarticletitle{Neural Fields in Visual Computing and Beyond}.
\newblock \bibinfo{journal}{\emph{Computer Graphics Forum}}
  (\bibinfo{year}{2022}).
\newblock
\showISSN{1467-8659}


\bibitem[\protect\citeauthoryear{Yang, Belongie, Hariharan, and Koltun}{Yang
  et~al\mbox{.}}{2021}]%
        {yang2021geometry}
\bibfield{author}{\bibinfo{person}{Guandao Yang}, \bibinfo{person}{Serge
  Belongie}, \bibinfo{person}{Bharath Hariharan}, {and}
  \bibinfo{person}{Vladlen Koltun}.} \bibinfo{year}{2021}\natexlab{}.
\newblock \showarticletitle{Geometry Processing with Neural Fields}.
\newblock \bibinfo{journal}{\emph{Advances in Neural Information Processing
  Systems}}  \bibinfo{volume}{34} (\bibinfo{year}{2021}).
\newblock


\bibitem[\protect\citeauthoryear{Yariv, Gu, Kasten, and Lipman}{Yariv
  et~al\mbox{.}}{2021}]%
        {Yariv2021}
\bibfield{author}{\bibinfo{person}{Lior Yariv}, \bibinfo{person}{Jiatao Gu},
  \bibinfo{person}{Yoni Kasten}, {and} \bibinfo{person}{Yaron Lipman}.}
  \bibinfo{year}{2021}\natexlab{}.
\newblock \showarticletitle{Volume Rendering of Neural Implicit Surfaces}.
\newblock \bibinfo{journal}{\emph{NeurIPS}} (\bibinfo{year}{2021}).
\newblock


\bibitem[\protect\citeauthoryear{Yifan, Rahmann, and Sorkine-hornung}{Yifan
  et~al\mbox{.}}{2022}]%
        {yifan2021geometry}
\bibfield{author}{\bibinfo{person}{Wang Yifan}, \bibinfo{person}{Lukas
  Rahmann}, {and} \bibinfo{person}{Olga Sorkine-hornung}.}
  \bibinfo{year}{2022}\natexlab{}.
\newblock \showarticletitle{Geometry-Consistent Neural Shape Representation
  with Implicit Displacement Fields}. In
  \bibinfo{booktitle}{\emph{International Conference on Learning
  Representations}}.
\newblock


\bibitem[\protect\citeauthoryear{Young}{Young}{1931}]%
        {young1931algebra}
\bibfield{author}{\bibinfo{person}{Rosalind~Cecily Young}.}
  \bibinfo{year}{1931}\natexlab{}.
\newblock \showarticletitle{The algebra of many-valued quantities}.
\newblock \bibinfo{journal}{\emph{Math. Ann.}} \bibinfo{volume}{104},
  \bibinfo{number}{1} (\bibinfo{year}{1931}), \bibinfo{pages}{260--290}.
\newblock


\end{thebibliography}

\appendix

\section{Affine Arithmetic Rules for MLPs}
\label{app:AffineArithmeticRules}

The operations needed to apply affine arithmetic to MLPs have been established in past work (\eg{} \citet{stol1997self}), but a variety of different conventions and notations may obscure implementation.
We gather these operations here in concise notation to facilitate future adoption.

\begin{table*}[]
\caption{The update rules to propagate affine arithmetic bounds on MLPs. All quantities are in-general vector or matrix-valued.}
\begin{tabular}{@{}rccccc@{}}
\toprule
  \textbf{Name}           
  & \textbf{Operation} 
  & \multicolumn{1}{l}{\textbf{New} $z_0$} 
  & \multicolumn{1}{l}{\quad \textbf{New} $Z$} 
  & \multicolumn{1}{l}{\textbf{New} $z_\infty$} 
  & \multicolumn{1}{l}{\textbf{Notes}} \\ \midrule
  constant addition       & $z \gets \affine{x} + a         $ & $x_0 + a$                 &  $X    $      & $ x_\infty$ & \\
  affine addition         & $z \gets \affine{x} + \affine{y}$ & $x_0 + y_0$               &  $X + Y$      & $ x_\infty + y_\infty$ & \\
  constant multiplication & $z \gets a\affine{x}            $ & $a x_0  $                 &  $aX $        & $ |a| x_\infty$   & \\
  matrix multiplication   & $z \gets A \affine{x}           $ & $A x_0  $                 &  $AX $        & $|A| x_\infty$    & $|A|$ is element-wise \\
  nonlinearities          & $z \gets h(\affine{x})          $ & $\alpha x_0 + \beta $     &  $[\alpha X; \textrm{diag}(\gamma)]$   & $|\alpha| x_\infty $ & see \eqref{affine_func_update} and \tabref{affine_nonlinearities} \\
\end{tabular}
\label{tab:affine_update_rules}
\end{table*}

\begin{table*}[]
\caption{
  Formulas for affine approximation of common nonlinear function in neural networks. To propagate bounds through each function, the parameters $\alpha$, $\beta$, $\gamma$ are computed for the given $\affine{x}$, and then applied to produce output $y$ as in \eqref{affine_func_update}.
  In implementation, care must be taken with fractional terms to ensure stability when $\intlower{x} = \intupper{x}$.
  Here, $\cos{([\intlower{x}, \intupper{x}])}$ denotes the maximum and minimum value of $\cos$ on $[\intlower{x}, \intupper{x}]$, which can be computed via modular arithmetic.
  The computed parameters for $\textrm{sin}(\affine{x})$ use the same form as Chebyshev approximations for convex functions, but are \emph{not} the Chebyshev approximation because the function is not convex.
  \label{tab:affine_nonlinearities}
}
\begin{tabular}{lllc}
\toprule
\multicolumn{1}{c}{\textbf{Nonlinearity}} & 
\multicolumn{1}{c}{\textbf{Parameters} $\alpha$,$\beta$,$\gamma$} & 
\multicolumn{1}{c}{\textbf{Notes}} & 
\textbf{Diagram} \\
\midrule
$\textrm{ReLU}(\affine{x})$ &
\begin{minipage}{0.3\textwidth}
\begin{equation*}
\begin{aligned}
  & [\intlower{x}, \intupper{x}] \gets \range{\affine{x}} \\
  & \alpha \gets \frac{\textrm{ReLU}(\intupper{x}) - \textrm{ReLU}(\intlower{x})}{\intupper{x} - \intlower{x}}\\
  & \beta \gets (\textrm{ReLU}(\intlower{x}) - \alpha \intlower{x}) / 2\\
  & \delta \gets \beta
\end{aligned}
\end{equation*}
\\
\end{minipage}
&   Chebyshev approximation   &
\adjustimage{height=3cm,valign=m}{images/affine_plot_relu}
\\
\midrule
$\textrm{ELU}(\affine{x})$ &
\begin{minipage}{0.3\textwidth}
\begin{equation*}
\begin{aligned}
  & [\intlower{x}, \intupper{x}] \gets \range{\affine{x}} \\
  & \textrm{if } \intlower{x_l} > 0:  \alpha \gets 1, \quad \beta \gets 0, \quad \gamma \gets 0 \\
  & \textrm{else: } \\
  & \alpha \gets \frac{\textrm{ELU}(\intupper{x}) - \textrm{ELU}(\intlower{x})}{\intupper{x} - \intlower{x}}\\
  & r_u \gets \textrm{ELU}(\intlower{x}) - \alpha \intlower{x} \\
  & r_l \gets (\alpha - 1.) - \alpha (\ln{\alpha} - \alpha \intlower{x}) \\
  & \beta \gets (r_u + r_l) / 2 \\
  & \delta \gets r_u - \beta \\
\end{aligned}
\end{equation*}
\end{minipage}
&   Chebyshev approximation   &
\adjustimage{height=3cm,valign=m}{images/affine_plot_elu} \\ 
\midrule
$\textrm{sin}(\affine{x})$ &
\begin{minipage}{0.3\textwidth}
\begin{equation*}
\begin{aligned}
  & [\intlower{x}, \intupper{x}] \gets \range{\affine{x}} \\
  & [\intlower{s}, \intupper{s}] \gets \cos{([\intlower{x}, \intupper{x}])} \\
  & \alpha \gets (\intlower{s} + \intupper{s}) / 2\\
  & e_p \gets \arccos{\alpha}, \quad e_n \gets -\arccos{\alpha} \\
  & \mathcal{E} \gets [ \intlower{x}, \intupper{x}, \\
  & \hspace{2.8em} \textrm{clamp}(2 \pi \textrm{ceil}(\intlower{x} + e_p) / (2 \pi) - e_p , \intlower{x}, \intupper{x}),  \\
  & \hspace{2.8em} \textrm{clamp}(2 \pi \textrm{ceil}(\intlower{x} + e_n) / (2 \pi) - e_n , \intlower{x}, \intupper{x}),  \\
  & \hspace{2.8em} \textrm{clamp}(2 \pi \textrm{floor}(\intlower{x} - e_p) / (2 \pi) + e_p, \intlower{x}, \intupper{x}),  \\
  & \hspace{2.8em} \textrm{clamp}(2 \pi \textrm{floor}(\intlower{x} - e_n) / (2 \pi) + e_n, \intlower{x}, \intupper{x}) \hspace{.2em} ] \\
  & r_u \gets \max_{e \in \mathcal{E}} (\sin(e) - \alpha e ) \\
  & r_l \gets \max_{e \in \mathcal{E}} (\sin(e) - \alpha e ) \\
  & \beta \gets (r_u + r_l) / 2 \\
  & \delta \gets r_u - \beta \\
\end{aligned}
\end{equation*}
\end{minipage}
&    &
\adjustimage{height=3cm,valign=m}{images/affine_plot_sin} \\ 
\bottomrule
\end{tabular}
\end{table*}

For any vector quantity $\vec{x} \in \mathbb{R}^m$ which arises while evaluating an MLP, we programmatically represent the affine approximation $\vec{\affine{x}}$ via the tuple of values $(\vec{x}_0, X, \vec{x}_\infty)$, where $\vec{x}_0 \in \mathbb{R}^m$ is the base value, $X \in \mathbb{R}^{m \times N}$ is a stacked matrix of $N$ column vectors encoding the coefficients for each affine term, and $\vec{x}_\infty \in \mathbb{R}^m$ is a special distinguished affine coefficient as a convenient notation to model terms truncated during condensation (\secref{ReducedAffineArithmetic}).
We distinguish the coefficient $x_\infty$ because terms varying due to $\varepsilon_\infty$ must be treated as distinct whenever they arise in expression---unlike other affine coefficients it does not \eg{} cancel under subtraction, because it captures variation from many distinct sources.
The vector-valued affine approximation of $x$ is then 
\begin{equation}
  \affine{x} = x_0 + \sum_{i < N} X_i \varepsilon_i + x_\infty \varepsilon_\infty
\end{equation}
where $X_i$ denotes the $i$'th column of $X$. 
For the remainder of this section, we will drop the vector notation and simply write $\affine{x}$ and $(x_0, X, x_\infty)$.

\setlength{\columnsep}{0.5em}
\setlength{\intextsep}{0em}
\begin{wrapfigure}{l}{42pt}
  \vspace{0.2em}
  \includegraphics{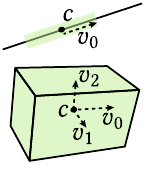}
\end{wrapfigure}
The affine representation is initialized as in \algref{RangeBound}: to construct an affine approximation of a not-necessarily axis-aligned $s$-dimensional box in $\mathbb{R}^d$, we take as input the center of the box $c \in \mathbb{R}^d$ and a collection of $s$ orthogonal vectors from the center to the sides of the box $\{v_0, ...,v_{s-1}\}$ (1d and 3d constructions shown inset).
The initial coefficients are then
\begin{equation*}
  x_0 \gets c, \qquad X \gets [v_0;...;v_s], \qquad x_\infty \gets \vec{0}.
\end{equation*}

To propagate these bounds forward through a network, we require rules to update bounds after addition (by a constant as well as other with affine quantities), multiplication by a constant, matrix multiplication by a constant, and activation functions.
\tabref{affine_update_rules} lists the rules, while \tabref{affine_nonlinearities} gives expressions for computing affine approximation parameters for common activation functions.

\section{Additional Details}
\label{app:AdditionalDetails}

Here, we give miscellaneous configuration details for the algorithms and experiments above. 

\paragraph{Range Analysis Empirical Study}

To construct a small benchmark dataset of neural implicit surfaces, we gather a collection of 10 shapes including characters, mechanical models, and 3D scans, and fit several MLPs to each, using all combinations of $\textrm{ReLU}$ \vs{} $\textrm{ELU}$ nonlinearities, as well as fitting SDFs under an $L_1$ penalty \vs{} occupancy under cross-entropy loss.
Each MLP has 7 hidden layers of width 32 for a total of 7553 parameters.
Training points are sampled as in \cite{Davies2020}, and we train for 100 epochs with the ADAM optimizer, using a batch size of 512 and learning rate of $10^{-2}$ decreased by a factor $10$ after 50 epochs.

Timings are measured in the experimental configuration described in \secref{Implementation}, and all time statistics are normalized to the fastest variant.
To quantify the tightness of the bounds from each range analysis variant, we sample many random regions of varying sizes, and test whether the implicit function can be bounded away from zero over the region.
Precisely, for any region size $s$, we compute $f$, the fraction of regions such that \algref{RangeBound} outputs \typepos{} or \typeneg{}. 
We then report the largest $s$ such that $f \geq 50\%$. 
Bigger $s$ means tighter bounds; this size is an indication of \eg{} how large of steps must be taken for ray casting, or how large the cells in the spatial hierarchy can be.
Values are averaged over 10,000 random regions per implicit surface, and for timings we furthermore take the fastest of 5 runs to account for warm-up and other variance.
We additionally measure the time to cast rays corresponding to pixels of a $256 \times 256$ camera view centered on the surface.

\paragraph{Ray Casting}

We march with a small safety tolerance, taking steps of size $0.98\sigma$ to mitigate floating point inaccuracy, although we do not observe failures due to floating point in any case.
For frustum ray casting, we initialize a $16 \times 16$ grid of coarse frusta, and subdivide frusta in half along the largest dimension whenever the width of that dimension along the forward face is greater than twice the current step size $\sigma$.
Additionally, we note that the length of the bounding box used for range analysis needs to be extended slightly beyond the front face of the frustum defined by the rays in its corners, because the contained rays actually sweep out a spherical region.
We compute the correct extent by forming a ray along the center of the frustum with length $t+\sigma$, and measuring the distance of its endpoint from the frustum base.

\paragraph{Mesh Extraction}

We build the tree with \texttt{affine-all} range analysis, and use dense
evaluation for the lowest $l=3$ levels.

\paragraph{Inverse Rendering}

The inverse rendering example in \figref{inverse_render_bunny} fits a $5$-layer, $128$-width network with ELU activations to 20 camera views, each at $512 \times 512$ resolution, equally spaced in a sphere around the subject.
The ground truth is rendered from a triangle mesh with Blinn-Phong shading by 3 fixed point lights in the scene.
The loss is $10\times$ the $L_1$ image difference on the rendered image, plus a cross-entropy occupancy loss on the minimum value of the implicit function as sampled at $100$ points along each ray.
Rays from each pixel in all views are combined and batched with size $512$. We train for 4 epochs using the ADAM optimizer, with a learning rate of $1e^{-3}$ decayed by a factor of $0.5$ on each epoch.
The rendered initial view in \figref{inverse_render_bunny} is shown after a small number (100) of training steps, because the randomly-initialized network on the first iteration is not visually coherent.

\end{document}